\documentclass[acmtog]{acmart}
\acmSubmissionID{1791}

\usepackage{booktabs} 

\usepackage{multirow}
\usepackage{subfigure}
\usepackage[normalem]{ulem}
\usepackage{pifont}
\usepackage{makecell}
\usepackage{float}

\usepackage[ruled]{algorithm2e} 

\SetAlFnt{\small}
\SetAlCapFnt{\small}
\SetAlCapNameFnt{\small}
\SetAlCapHSkip{0pt}

\acmJournal{TOG}

\setcopyright{cc}
\setcctype{by}
\acmJournal{TOG}
\acmYear{2026} \acmVolume{45} \acmNumber{6} \acmArticle{188}
\acmMonth{12} \acmDOI{10.1145/3842544}

\newcommand{\xin}[1]{{\color{cyan}{#1}}}

\author{Haotian Dong}
\orcid{0009-0006-7137-6955}
\affiliation{%
 \institution{Tianjin University}
 \city{Tianjin}
 \country{China}}
\email{htdong@tju.edu.cn}

\author{Wenjing Wang}
\orcid{0000-0003-3951-3877}
\affiliation{%
 \institution{Independent}
 \city{Beijing}
 \country{China}}
\email{augustawang@tencent.com}

\author{Chen Li}
\orcid{0000-0002-2450-8525}
\affiliation{%
 \institution{Independent}
 \city{Beijing}
 \country{China}}
\email{chaselli@tencent.com}

\author{Jing Lyu}
\orcid{0009-0004-2021-0256}
\affiliation{%
 \institution{Independent}
 \city{Beijing}
 \country{China}}
\email{eckolv@tencent.com}

\author{Xin Wang}
\orcid{0000-0002-7977-6586}
\authornote{Co-corresponding author.}
\affiliation{%
 \institution{The Hong Kong Polytechnic University}
 \country{Hong Kong}}
\email{xin1025.wang@connect.polyu.hk}

\author{Di Lin}
\orcid{0000-0002-9324-800X}
\authornotemark[1]
\affiliation{%
 \institution{Tianjin University}
 \city{Tianjin}
 \country{China}}
\email{ande.lin1988@gmail.com}

\begin{document}
\title{Loopy: Seamless Video Loop Generation via Anchored Looping Shift of Positional Embedding}

\begin{abstract}
Looping videos are essential for practical applications such as web graphics, game development, and social media.
However, existing approaches typically fail to generate high-quality looping videos due to the neglect of how video generation models perceive temporal order and how this relates to the looping behavior.
In this work, we are the first to reveal that position embedding at different attention layers within DiT
exhibits varying levels of positional control, with the most 
pronounced layer acting as an anchor. We formulate this anchored layer as the reference point of the looping video, offering strong contextual priors for the remaining layers to facilitate the generation of seamless and coherent video content.
Based on this insight, we propose an anchored position embedding shifting strategy that applies layer-specific shift lengths according to each layer’s temporal control 
effect, effectively transforming DiT’s temporal perception from a straight line to a circle.
Leveraging this strategy, we develop a general framework, Loopy, for high-quality looping video generation, supporting both RGB and RGBA videos, while also enabling advanced AIGC features such as identity control and style transfer.
Experiments demonstrate that our approach significantly improves temporal consistency and visual fidelity in generated looping videos. {The released model is available on our
website: https://donghaotian123.github.io/Loopy.}

\end{abstract}

%
%


%


\begin{teaserfigure}
    \centering
    \includegraphics[width=0.99\linewidth]{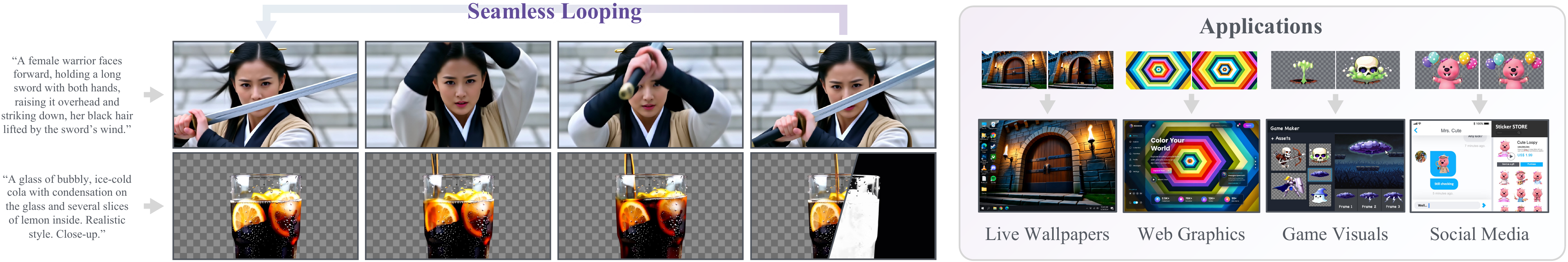}
    \vspace{-1mm}
    \caption{Our \textbf{Loopy} generates high-quality looping videos with seamless transitions at loop boundaries and diverse motion. It also supports RGBA with semi-transparent effects. In the application block, all elements—including game assets and Loopy character stickers—are generated by our \textbf{Loopy}.
    }
    \label{fig:teaser}
\end{teaserfigure}

\maketitle
\section{Introduction}
Seamless looping videos create the illusion of infinite duration and are widely used in applications such as dynamic wallpapers, web graphics, advertising design, and visual effects.
Despite the high demand, producing high-quality looping videos is costly and labor-intensive.
This motivates the use of Artificial Intelligence-Generated Content (AIGC) algorithms as a natural and promising approach for looping video generation.
The community has witnessed substantial progress in video generation technologies~\cite{wan2025wan,hunyuanvideo_1.5,klingteam2025klingomnitechnicalreport,seedance2026seedance20advancingvideo}, which allow users to control generated content through text prompts.

\begin{figure}[t!]
    \centering
    \vspace{-1mm}
    \includegraphics[width=0.99\linewidth]{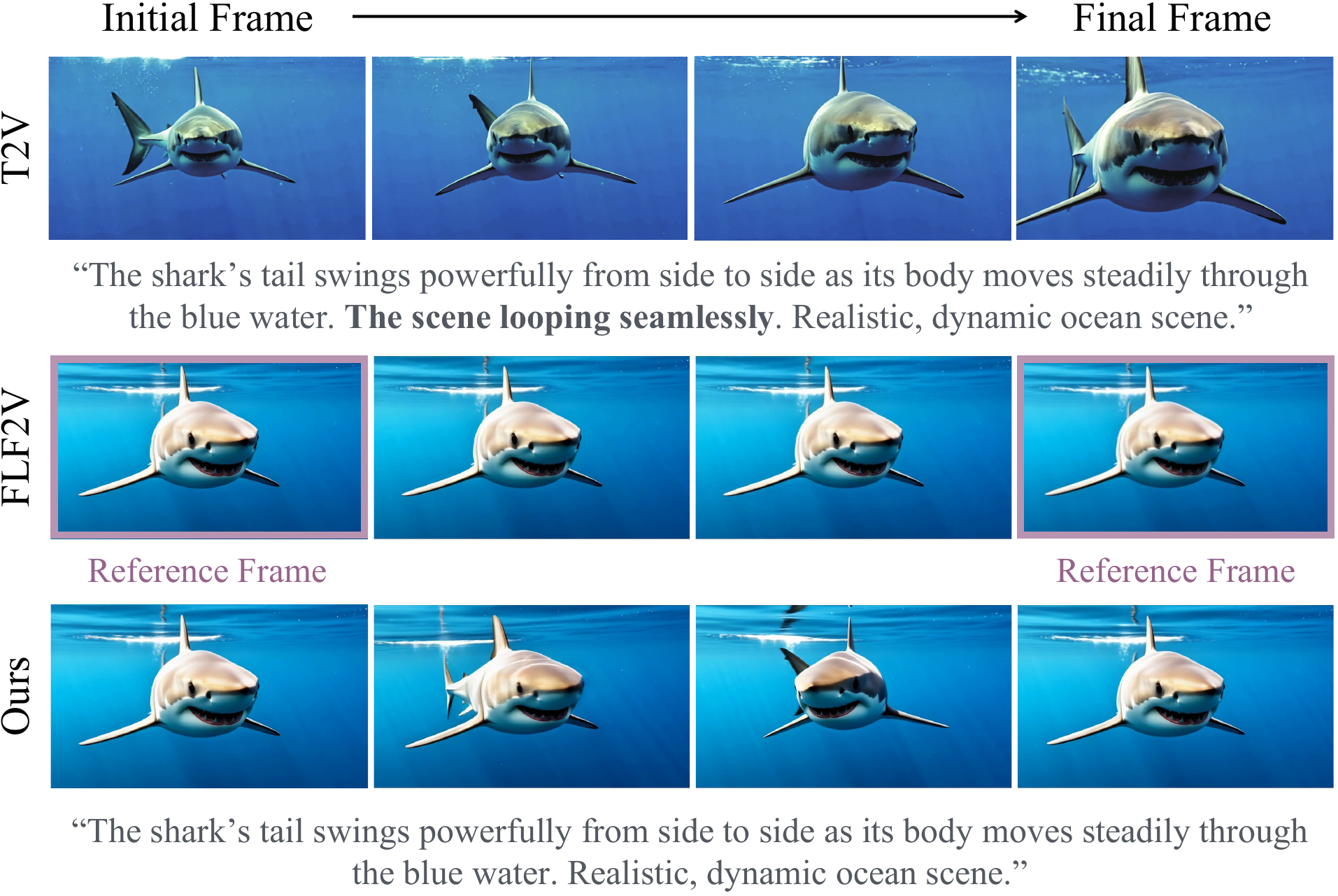}
    \vspace{-1mm}
    \caption{{Comparison with Wan2.2~\cite{wan2025wan} in text-to-video (T2V) and first-last-frame-to-video (FLF2V) modes. FLF2V and our method use the same prompt. Although the prompt explicitly requests a looping scene, T2V fails to produce a seamless loop, while FLF2V degenerates to a nearly static video. Neither mode generates a looping video with vivid motion.}}
    \label{fig:intro_t2v_flf2v}
    \vspace{-1mm}
\end{figure}

However, these models are not explicitly designed for looping video generation, limiting their direct use in seamless loop generation from text prompts, as shown in the first row of Fig.~\ref{fig:intro_t2v_flf2v}.
Fine-tuning existing video generation models on looping data is an intuitive solution, but the narrow access to a large volume of such data limits the effectiveness of the fine-tuning strategy.
Another straightforward solution is to utilize first-last-frame-to-video generation models~\cite{wan2025wan} and enforce the first and last frames to be the same~\cite{mahapatra2026dreamloop}.
However, such models tend to converge toward the path of least resistance and suffer from static collapse, leading to limited motion variation or even fully static videos as shown in the second row of Fig.~\ref{fig:intro_t2v_flf2v}.
Consequently, existing general video generation models struggle to generate seamless looping videos with diverse yet coherent motions.

To handle looping, training-free strategies~\cite{bi2025mobius,latentmix} have recently emerged by manipulating latent representations during the diffusion process.
However, 
such simple latent manipulations neglect the co-effects of the video generation model, often introducing unnatural motion variations and temporal inconsistencies.
To provide looping videos for network training, LoopAnimate~\cite{wang2024loopanimate} introduces an asymmetric loop sampling strategy, which involves sampling frames first forward and then backward with a sequence of particular sampling steps. However, such data generation strategy inevitably disrupts real-world motion patterns, limiting the diversity of motion and scenes and restricting their applicability primarily to scenarios with salient foreground objects. 
Moreover, existing methods neglect how video generation models perceive temporal order and how this relates to the formation of coherent looping behavior.

The core challenge lies in establishing a temporal 
coherence between the last and first frames with limited high-quality looping video data.
In this paper, we propose a novel framework for seamless looping video generation based on a thorough analysis on how video generation models perceive temporal order. 
State-of-the-art video generation models~\cite{wan2025wan,hunyuanvideo_1.5} typically adopt the Diffusion Transformer (DiT)~\cite{peebles2023scalable} backbone, which perceives spatiotemporal location using Rotary Position Embedding (RoPE)~\cite{su2024roformer} within attention layers.
Since the temporal encoding in RoPE progresses from the first frame to the last, such linearity prevents video generation models from establishing temporal continuity between the final and initial frames.
Moreover, RoPE is typically applied to each attention layer, but the differences in its temporal effect across layers have not been explored.

To the best of our knowledge, this work is the first to analyze RoPE's control over temporal position at different attention layers within DiT. 
{We observe that naively shifting all RoPEs with progressively increasing offsets encourages the DiT to perceive temporal positions cyclically. However, this simple RoPE-shifting strategy struggles to achieve seamless looping, as \textit{different RoPEs exhibit distinct degrees of control over temporal perception}.}
Through a quantitative analysis of such temporal control effects,
we further observe that \textit{the attention layer exhibiting the most pronounced RoPE control effect in DiT acts as an anchor}.
{This anchored layer provides critical contextual priors that shape the generated video content and reduce artifacts, while RoPEs in the remaining layers adjust their temporal effects relative to this reference.}

Motivated by these findings, we propose \textit{Anchored Position Embedding Shifting} (Anchored Shifting) to preserve temporal coherence and ensure seamless loop-boundary transition. Specifically, we shift RoPE within each attention layer along the temporal dimension by layer-specific shifting lengths, where the length is determined by each layer's degree of temporal control effect.
Our Anchored Shifting strategy enables the timeline perceived by DiT to change from a straight line to a {uniform} circle, strengthening the correlation between the final and initial frames.

Based on Anchored Shifting, we develop \textbf{Loopy}, a novel framework for looping video generation.
We first apply our Anchored Shifting to the most recent open-source video generation backbone, Wan 2.2 14B~\cite{wan2025wan}, to collect high-quality looping videos.
Then, we perform minimal fine-tuning on target video generation models using these high-quality looping video data and our Anchored Shifting strategy, enhancing seamless looping and improving coherence and diversity of motion. 
As shown in Fig.~\ref{fig:teaser}, our \textbf{Loopy} supports both RGB and more challenging RGBA looping video generation. The inclusion of a transparency channel enables flexible content manipulation, making RGBA looping videos particularly 
valuable for practical content creation and editing in applications such as game development, video effects, and digital design.
By integrating RGB and RGBA, our framework can also generate multi-layer looping videos.
Moreover, since our framework largely preserves the original architecture of video generation models, it can be jointly used with other AIGC features, such as controlling style and character in looping videos.
In summary:

\begin{itemize}
\item We are the first to reveal that RoPE in different DiT blocks exhibits varying levels of positional control, and that the most influential layer acts as an anchor.
\item We propose an anchored position embedding shifting strategy that applies a layer-specific shifting length based on each layer’s temporal control effect, enabling DiT’s temporal perception to change from a straight line to a circle.
On this basis, we construct a high-quality looping video dataset and train video generation models with {our shifting strategy}.
\item We develop a general framework for looping video generation, enabling high-quality RGB, RGBA, and multi-layer looping video generation.
\textbf{Loopy}
also supports other AIGC features, such as character control and style control.
\end{itemize}

\section{Related Work}
Our method is related to existing work on video generation and seamless looping video generation. Here, we primarily discuss works that are close to our Loopy.

\subsection{Video Generation}
Visual content generation, {including} both RGB~\cite{HunyuanVideo,wan2025wan,wang2025hrc,open_sora_2,10377659,Align_Your_Latents,11444877,show_1,seedance15pronative,10.1145/3757377.3763929,hunyuanvideo_1.5,wu2026x2hdr,klingteam2025klingomnitechnicalreport,seedance2026seedance20advancingvideo} and RGBA~\cite{chen2025transanimate,TransVDM,layeranimate,ILDiff,dong2025wanalpha,wang2025transpixeler} modalities, remains a challenging task that has {received} substantial attention in recent years.
Recent state-of-the-art video generation models typically adopt {the} DiT architecture due to its superior scalability and generation quality.
For instance, \citet{HunyuanVideo} {scale} the DiT backbone and {introduce} a unified dual-stream-to-single-stream architecture, {enabling} video generation with high visual quality, diverse motion, and {stability}.
\citet{open_sora_2} first {demonstrate} the potential of DiT-based video generation, producing minute-long, photorealistic clips with strong temporal consistency.
\citet{wan2025wan} {couple} a dedicated spatio-temporal VAE with a DiT denoiser trained via flow matching, achieving leading performance on both text-to-video and image-to-video tasks.
As an important subfield, RGBA video generation focuses on {producing} videos that {include} both RGB and transparency channels.
Existing methods~\cite{chen2025transanimate,TransVDM,layeranimate,ILDiff} {attempt to combine} the RGBA image generation framework, LayerDiffuse~\cite{LayerDiffuse}, into video generation models for RGBA video generation.
However, this straightforward integration causes temporal inconsistencies.
To address this limitation, recent works~\cite{dong2025wanalpha,wang2025transpixeler} fine-tune video generation models for RGBA video generation.
Specifically, \citet{wang2025transpixeler} {duplicate tokens} to jointly generate RGB and transparent videos.
\citet{dong2025wanalpha} {propose} a shiftable RGBA distribution learner, enabling high-quality transparent video generation.
{These approaches, whether RGB or RGBA, struggle to generate seamless looping videos, as they train on linear video clips and neglect the temporal coherence and contextual features inherent to looping videos.
Our Loopy addresses these limitations by introducing layer-specific temporal shifts to RoPE, transforming the temporal perception in DiT from a linear sequence into a cyclic structure. This design effectively improves temporal coherence between the final and initial frames, enabling seamless looping video generation.}

\subsection{Seamless Looping Video Generation}
{{Before the emergence of deep generative models, conventional pre-AI methods primarily formulated looping video generation as a content reuse and resampling problem, where existing visual frames were rearranged or reused to create video loops.}
\citet{Schodl} first propose this task by identifying similar frames as transition points and re-sequencing the video into a looping video. 
\citet{Kwatra} improve transition quality through spatiotemporal graph cuts, which find low-cost seams between video clips at the pixel level. \citet{Agarwala} extend this framework to create a panoramic video texture. \citet{liao2013automated} further introduce a spatially varying looping period, which represents varying levels of dynamism. Despite advances in seam-handling techniques, these methods remain fundamentally constrained by the input videos. They often struggle to generate large-scale motions and diverse dynamic visual patterns, since they primarily rely on rearranging or reusing existing video frames.
In contrast, our method directly generates text-conditioned looping videos, jointly modeling semantic content, motion, and loop closure without requiring a pre-existing video.

{Deep generative models have emerged as a promising solution for looping video generation, enabling the generation of novel temporal content beyond the constraints of existing input videos.}}
Several approaches~\cite{halperin2021endless,mahapatra2023text,bertiche2023blowing,mahapatra2026dreamloop} {treat} seamless looping video generation as {a form of cinemagraph creation.
They assume that the background is static and restrict motion to a specific mask, handling only scenarios with limited scene variation.}
To address this limitation, recent works~\cite{bi2025mobius,wang2024loopanimate} explore seamless looping video generation.
Specifically, \citet{bi2025mobius} {propose} a training-free method that rotates the latent at each denoising step to mitigate the scarcity of looping video data; however, artifacts may arise when this inference strategy does not suit all base video models. In contrast, \citet{wang2024loopanimate} {propose} LoopAnimate, which incorporates multi-stage condition initialization and a multi-level appearance and textual semantic decoupling module to balance motion variation and consistency in generated looping RGB videos.
Nonetheless, videos generated by LoopAnimate often exhibit limited motion variation due to insufficient modeling of the temporal and contextual coherence inherent in looping videos.
Furthermore, the methods above are designed for RGB looping videos, and directly applying {them} to RGBA produces unsatisfactory results. In contrast, our framework achieves seamless looping {for both RGB and RGBA videos}.

\section{Method}

\subsection{Preliminaries}
\label{sec:preliminary}

{Diffusion models~\cite{ho2020denoising} gradually transform data into noise through a Markovian forward diffusion process. 
By reversing this process, a model can generate realistic samples from noise. 
Flow Matching~\cite{lipman2023flowmatching,sit,sd3} generalizes this idea to continuous time: instead of a discrete sequence of noisy steps, it defines a time-continuous flow that directly transforms a simple base distribution $\mathcal{N}(0,I)$ into the data distribution $p_{\text{data}}$, allowing for flexible and efficient sample generation. The transformation is denoted as:
\begin{equation}
\begin{aligned}
    z_n &= (1-n) z + n \epsilon, \\
    z_0 &=z \sim p_{\text{data}}, \; z_1=\epsilon \sim \mathcal{N}(0,I), \quad n \in [0,1].
\end{aligned}
\end{equation}
In actual training, $n$ typically takes the form of a discrete time sequence $\{0, \Delta n, \dots, 1-\Delta n, 1\}$.
The network is trained to learn a vector field $\hat{v}_n$ that can transform $\epsilon$ into $z$:
\begin{equation}
\mathcal{L}_{\text{flow}} = \mathbb{E}_{z, \epsilon, n} \Big[ \big\| \hat{v}_n(z_n, n) - v_n \big\|^2 \Big], \quad v_n = \epsilon - z.
\end{equation}
After training, given $\epsilon = z_1$, the model can recover $z$ via:
\begin{equation}
    z_{n-\Delta n} = z_n - \hat{v}_n(z_n, n) \, \Delta n, \quad n = 1, 1-\Delta n, \dots, 0.
\end{equation}
The number of diffusion steps is defined as $N = \frac{1}{\Delta n}$.}

{Diffusion Transformer (DiT)~\cite{peebles2023scalable} has recently been widely used for large-scale image and video generation.
Compared with traditional U-Nets~\cite{unet}, DiT inherits the favorable scaling properties of transformers~\cite{transformer_2017}, with generation quality consistently improving as model depth, width, and training compute increase.
First, DiT divides the noisy input $z_n$ into non-overlapping patches and projects each patch into a token embedding.
The token embeddings are subsequently processed by a stack of transformer blocks, each composed of multi-head self-attention and a feed-forward network.
The diffusion timestep $n$ and conditioning signals, such as class labels or text embeddings, are injected into the transformer blocks through adaptive layer normalization or cross-attention mechanisms.
After the final block, the tokens are linearly projected and unpatchified back to the original shape to predict the velocity target $\hat{v}_n$.}

{Following LDM~\cite{ldm}, a pretrained Variational Autoencoder (VAE) is commonly employed to encode image or video pixels into latent representations, reducing computational cost. In the video domain, a 3D VAE compresses a $T$-frame input video $x \in \mathbb{R}^{T \times 3 \times H \times W}$ into a compact latent tensor $z \in \mathbb{R}^{T' \times C \times H' \times W'}$, which is subsequently patchified by DiT into a 3D token sequence. To capture long-range dependencies across frames, recent video DiTs~\cite{wan2025wan,hunyuanvideo_1.5} adopt full 3D self-attention over all spatio-temporal tokens, augmented with 3D Rotary Position Embedding (RoPE)~\cite{su2024roformer} for relative position encoding. Text conditioning is typically incorporated via cross-attention or a dual-stream design, where text and video tokens are jointly attended within the same transformer block. This unified token-based formulation enables DiT to flexibly handle variable resolutions, durations, and aspect ratios, providing a robust and scalable backbone for our method.}

\begin{figure*}[t!]
    \centering
    \includegraphics[width=0.99\linewidth]{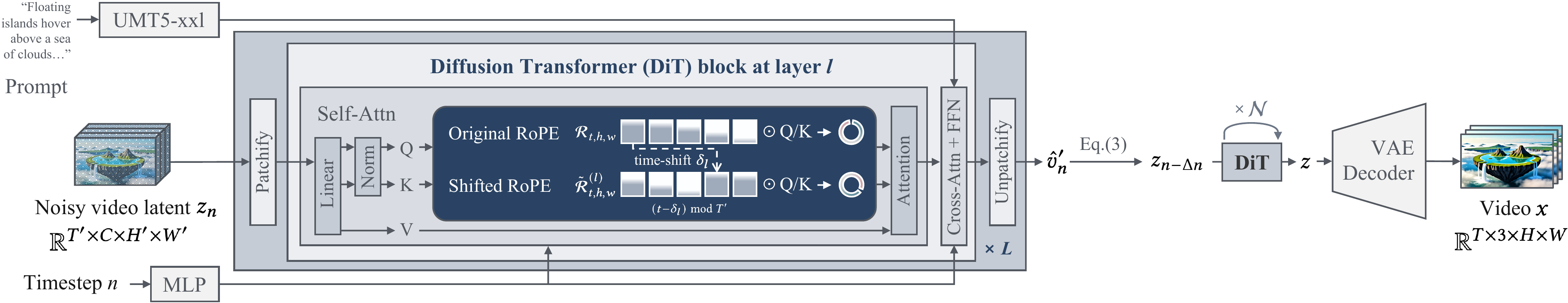}
    \vspace{-1mm}
    \caption{{The proposed temporal shifting operation on Wan2.2 14B~\cite{wan2025wan}. Given a noisy video latent $z_n$, DiT first patchifies it and maps patches into tokens, which are processed by $L$ DiT blocks. At layer $l$, self-attention applies a temporal shifting offset $\delta_l$ to the RoPE, which is then applied to the Q and K matrices. After all DiT blocks, tokens are unpatchified back to latents to produce $\hat{v}'_n$. Iterative diffusion denoising yields the predicted clean latent $z$, which is decoded by the VAE to generate the final video.}}
    \label{fig:framework_shift}
\end{figure*}

\begin{figure*}[t!]
    \centering
    \includegraphics[width=0.99\linewidth]{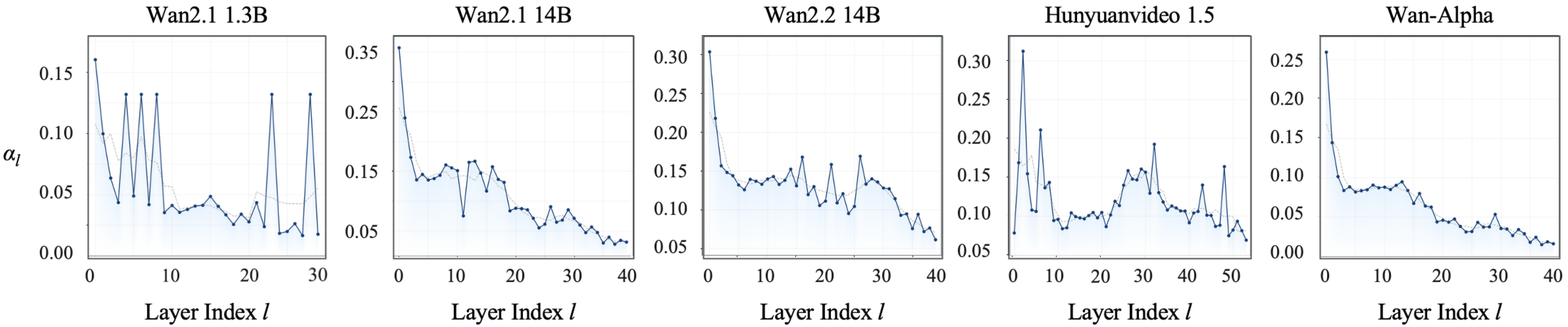}
    \vspace{-1mm}
    \caption{{Effect of RoPE on the temporal positional control $\alpha_l$ at the $l^{\text{th}}$ attention layer across state-of-the-art video generation models. Wan2.1 1.3B~\cite{wan2025wan,hunyuanvideo_1.5}, Wan2.1 14B~\cite{wan2025wan,hunyuanvideo_1.5}, Wan2.2 14B~\cite{wan2025wan,hunyuanvideo_1.5}, and Hunyuanvideo 1.5~\cite{hunyuanvideo_1.5} are used for RGB looping video generation, while Wan-Alpha~\cite{dong2025wanalpha} is used for RGBA looping video generation.
    }}
    \label{fig:analysis}
    \vspace{-1mm}
\end{figure*}

\begin{figure}[t!]
    \centering
    \vspace{-1mm}
    \includegraphics[width=0.99\linewidth]{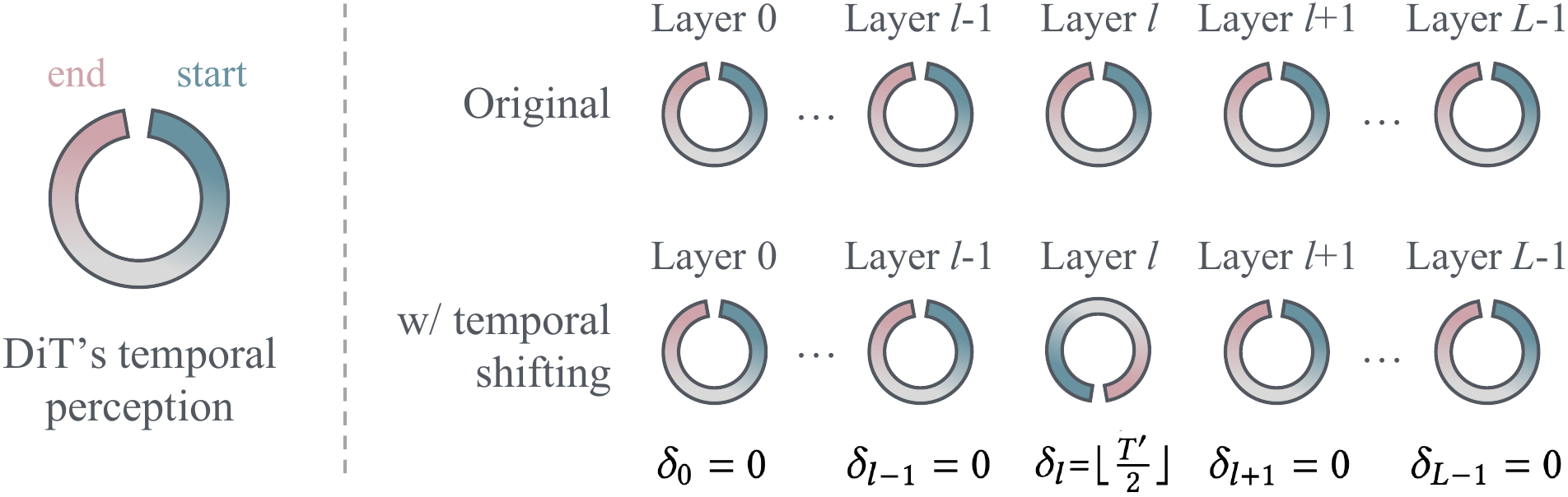}
    \vspace{-1mm}
    \caption{{Comparison between the original mode and our temporal shifting. The open ring visualizes DiT's temporal perception, with clockwise motion indicating video progression. Continuous segments represent periods where DiT preserves temporal continuity, while gaps mark discontinuities—specifically, the transition from the video's end back to its beginning.}}
    \label{fig:illus_shift}
    \vspace{-1mm}
\end{figure}

\begin{figure*}[t!]
    \centering
    \vspace{-1mm}
    \includegraphics[width=0.99\linewidth]{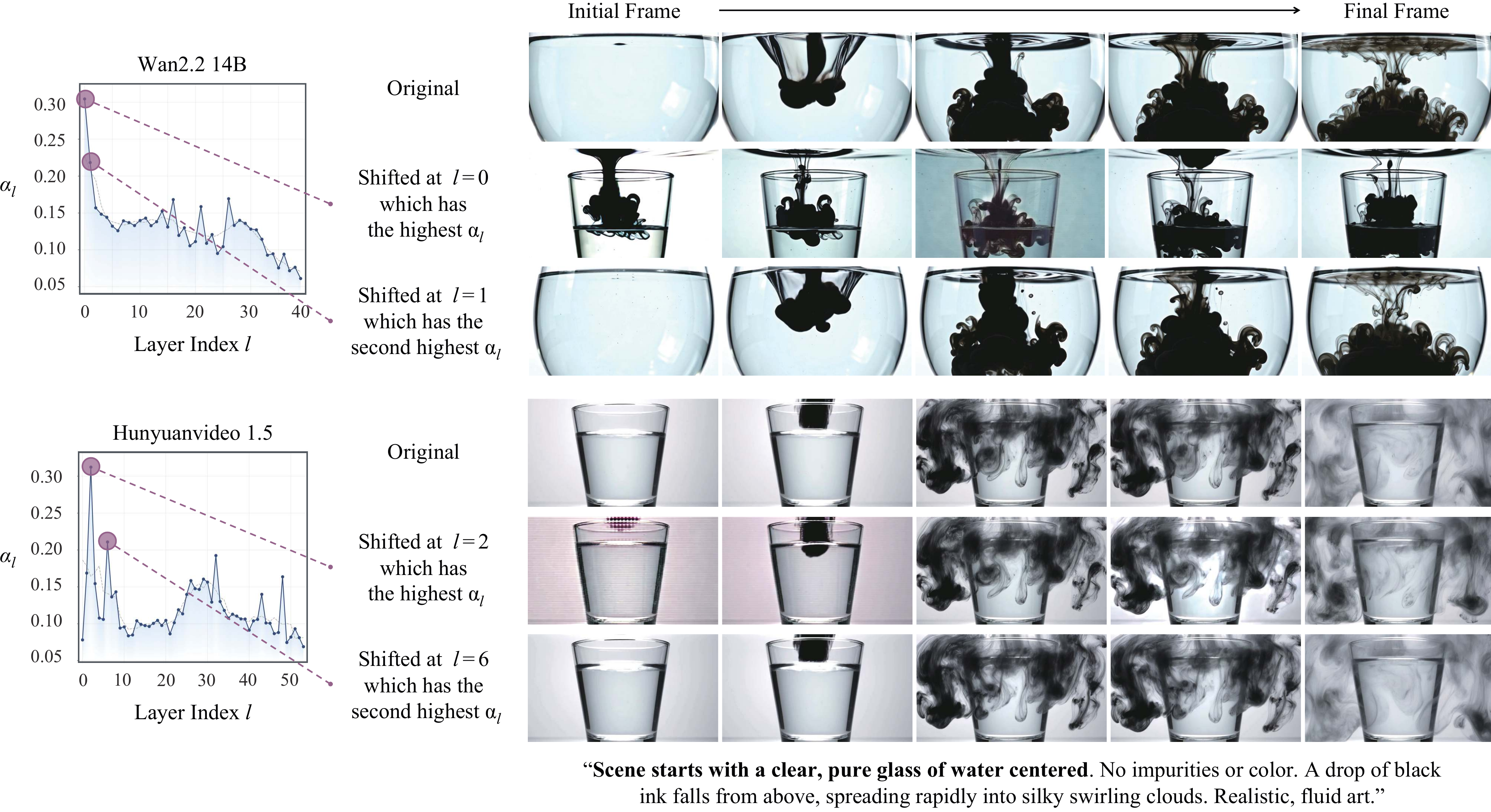}
    \vspace{-1mm}
    \caption{{Comparison of Wan2.2 14B~\cite{wan2025wan} and HunyuanVideo 1.5~\cite{hunyuanvideo_1.5} outputs when shifting RoPE at the layers with the first- and second-largest $\alpha_l$, relative to the original generated videos.}}
    \label{fig:anchor_result}
    \vspace{-1mm}
\end{figure*}

\subsection{Layer-wise Position Embedding Dependency}

\label{sec:dependency}
{As introduced in Section~\ref{sec:preliminary}, DiT includes many attention layers and typically uses RoPE to capture positional information. RoPE encodes positions linearly along the temporal, height, and width dimensions, with the temporal dimension progressing sequentially from the initial to the final video frame.
This design enables video generation models to learn the association between positional encoding and spatiotemporal consistency through training on large-scale data. However, the lack of inherent looping consistency from the final to the initial frame in typical video data prevents existing models from producing looping videos.}

{We observe that RoPE exhibits different levels of temporal positional control across attention layers in DiT. To analyze this layer-specific temporal control, we design a quantitative evaluation scheme.}
First, we define a temporal shifting operation of RoPE.
Standard 3D RoPE~\cite{su2024roformer} encodes position information by rotating query and key vectors in the complex plane ${\bf x}$. 
This rotation utilizes fixed rotary embeddings ${\mathcal{R}}$ that apply a rotation operator $e^{i\theta}$ for each dimension.
The rotation operator $e^{i\theta}$ is formulated as:
\begin{equation}
    e^{i\theta} = \cos\theta + i\cdot\sin\theta,
\end{equation}
where $\theta$ denotes the rotary angle or the base frequency.
{In video generation, the rotation to ${\bf x}$ is applied along three dimensions: temporal position $t$, and spatial positions $h$ (height) and $w$ (width). 
For a position $(t, h, w)$, the corresponding rotary embeddings are computed as:
\begin{equation}
\begin{aligned}
    &\boldsymbol{\mathcal{R}}_t = e^{i t \boldsymbol{\theta}^{D_t}}, \; \boldsymbol{\mathcal{R}}_h = e^{i h \boldsymbol{\theta}^{D_h}},\; \boldsymbol{\mathcal{R}}_w = e^{i w \boldsymbol{\theta}^{D_w}}, \\
    &\boldsymbol{\mathcal{R}}_{t,h,w} = \left[ \boldsymbol{\mathcal{R}}_t, \boldsymbol{\mathcal{R}}_h, \boldsymbol{\mathcal{R}}_w \right] ,\\
\end{aligned}
\end{equation}
where $\boldsymbol{\mathcal{R}}_{*}$ denotes the rotary embedding for position, and $\boldsymbol{\theta}^{D_*}$ represents the base frequencies for the corresponding dimension.}
For the $l^{th}$ layer, we formulate the modified rotary embeddings with a layer-specific time-shift {offset} $\delta_l$ as:
\begin{equation}
    \tilde{\boldsymbol{\mathcal{R}}}^{(l)}_{t,h,w} = \boldsymbol{\mathcal{R}}_{(t - \delta_l) \bmod T', \, h, \, w}~,
\label{eq:shift_rope}
\end{equation}
where $l\in \{0,\cdots,L-1\}$ is the index of layer, and $T'$ indicates the length of the latent representation corresponding to the input $T'$-frame video.
Finally, we use the modified rotary embeddings $\tilde{\boldsymbol{\mathcal{R}}}$ to shift the complex plane ${\bf x}$, yielding rotated query and key vectors in the complex plane ${\bf x}'$ as:
\begin{equation}
    {\bf x}'_{t,h,w} = \tilde{\boldsymbol{\mathcal{R}}}^{(l)}_{t,h,w} \odot \mathbf{x}_{t,h,w}.
\label{eq:shift_x}
\end{equation}
{where $\odot$ denotes element-wise multiplication. An illustration of the proposed temporal shifting operation on Wan2.2 14B~\cite{wan2025wan} is shown in Fig.~\ref{fig:framework_shift}.}

To quantitatively evaluate the degree of temporal positional control affected by RoPE within the $l^{th}$ attention layer, {we set the time-shift offset to} $\delta_l$=$ \lfloor \frac{T'}{2} \rfloor$, enabling the RoPE to be shifted by half the video length along the temporal dimension. {For all other layers, we set the time-shift offset to $0$. An illustration is shown in Fig.~\ref{fig:illus_shift}.}
We then utilize normalized Mean Squared Error (MSE) to quantify the extent to which such shifting alters the predictions of the DiT. Given the output of the original DiT $\hat{v}_n$ and the corresponding output after time shift $\hat{v}'_n$ at the $n^{th}$ diffusion timestep, we formulate the calculation of the average normalized MSE $\alpha_l$ over all diffusion timesteps $N$ as:
\begin{equation}
\alpha_l = \mathbb{E}_{n \sim N} \left[ \left\| \hat{v}'_n - \hat{v}_n \right\|_2^2 \right]
\end{equation}
{To reduce content-specific variations, we further averaged $\alpha_l$ over 100 prompts spanning diverse scenes.}

{In Fig.~\ref{fig:analysis}, we show results for the most representative open-source text-to-video models currently available, including both RGB~\cite{wan2025wan,hunyuanvideo_1.5} and RGBA~\cite{dong2025wanalpha}. Applying a $ \lfloor \frac{T'}{2} \rfloor$-shifted RoPE at different attention layers introduces varying levels of bias in DiT outputs. A consistent trend is that earlier layers have a stronger effect than later layers, which matches empirical observations that DiT’s early layers primarily capture structural information. Across all models, certain layers stand out with significantly higher influence, such as the first layer in Wan2.2 14B and the third layer in HunyuanVideo 1.5. We discuss these patterns in the next section.}

\subsection{Anchor as Contextual Prior}
\label{sec:anchor}
Looping video generation involves two fundamental objectives: looping consistency and semantic fidelity. Existing video generation models primarily focus on looping consistency, which aims to ensure temporal coherence between the final and initial frames, thereby enabling seamless transitions at looping boundaries. In contrast, semantic fidelity has received comparatively limited attention. Semantic fidelity requires the generated looping video to remain consistent with the user-specific semantics throughout the entire video sequence, ensuring that the visual content continuously aligns with the input text prompt as the video progresses from the initial frame to the final.

At the architecture level of existing state-of-the-art video generation models, simultaneously ensuring looping consistency and semantic fidelity is challenging, as these two objectives are inherently at odds with each other. Text and video modalities can be integrated through {self-attention~\cite{Phenaki} or cross-attention~\cite{wan2025wan}} mechanisms to enhance semantic fidelity in video generation. However, this design tends to preserve the linear temporal ordering learned from pretrained models, thereby hindering the generation of seamless looping videos. {In contrast, achieving looping consistency requires constructing a new temporal ordering.}

{In Section~\ref{sec:dependency}, we observed that shifting RoPE at different attention layers introduces varying degrees of bias into DiT outputs. This effect also appears in semantic fidelity, with the attention layer exhibiting the strongest RoPE control playing a distinct role compared to other layers. Videos generated with RoPE shifts at different layers are shown in Fig.~\ref{fig:anchor_result}.
For a prompt requiring the video to start with a clear, pure glass of water, Wan2.2 14B produces a video starting with black ink when RoPE is shifted at layer $0$, which has the highest $\alpha_l$. This occurs because shifting RoPE by $\delta_0 = \lfloor \frac{T'}{2} \rfloor$ maps tokens from the initial to final frames to new temporal positions $\{\lfloor \frac{T'}{2} \rfloor, \lfloor \frac{T'}{2} \rfloor+1, \dots, T'-1, 0, 1, \dots, \lfloor \frac{T'}{2} \rfloor-1\}$, effectively placing the initial frame in the middle of the sequence. Other layers ($l>0$) also affect semantic fidelity, but to a lesser extent.  
Although shifting RoPE at layer $0$ introduces a temporal discontinuity at the middle of the sequence (between $T'-1$ and $0$), the motion remains largely smooth due to layers $l>0$, whose RoPE is unshifted and preserve temporal consistency. Nevertheless, a dark green bias appears in the middle of the video, demonstrating the dominant influence of layer $0$.  
For HunyuanVideo 1.5, shifting RoPE at layer $2$ produces a noticeable red color bias and checkerboard artifacts, indicating that layer $2$ has a stronger influence than the other layers.
In contrast, for both models, shifting the layer with the second-highest $\alpha_l$ produces no semantic changes or color bias, indicating that it has much less influence than the layer with the highest $\alpha_l$.}

{We refer to the attention layer with the strongest RoPE control as the anchor layer, which serves as a contextual prior for video generation. Shifting the RoPE of the anchor layer may introduce undesirable artifacts or semantic inconsistencies. By keeping the anchor layer unchanged, semantic fidelity is better preserved and artifacts are greatly reduced.}

\subsection{Anchored Position Embedding Shifting}

\begin{figure}[t!]
    \centering
    \vspace{-1mm}
    \includegraphics[width=0.99\linewidth]{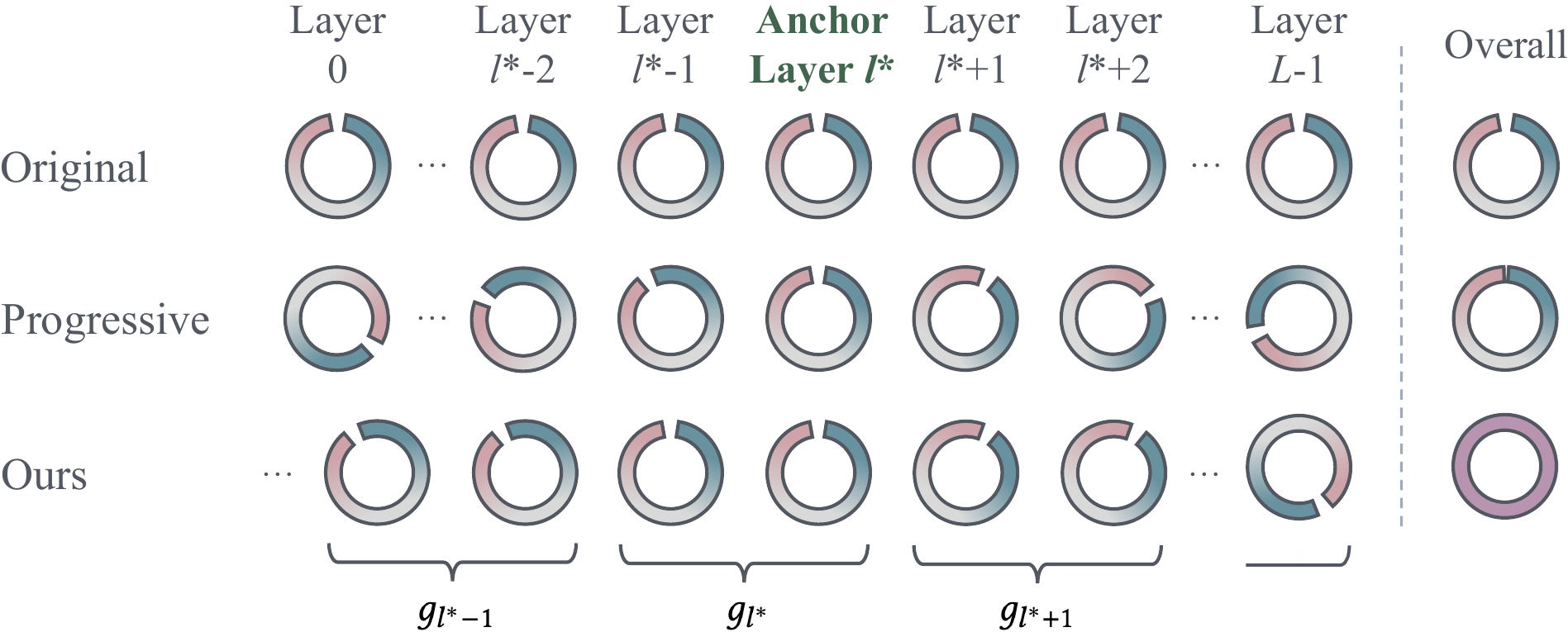}
    \vspace{-1mm}
    \caption{{Comparison of the original mode, the progressively increasing strategy, and our method.}}
    \label{fig:shifting_strategy}
    \vspace{-1mm}
\end{figure}

\begin{figure}[t!]
    \centering
    \vspace{-1mm}
    \includegraphics[width=0.99\linewidth]{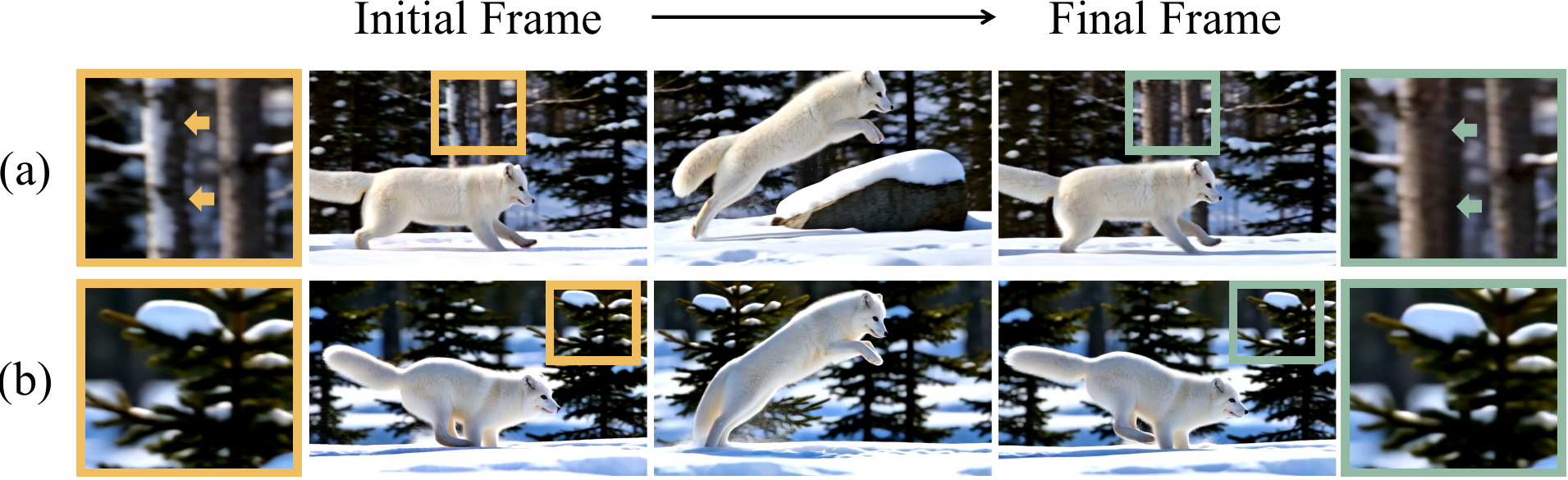}
    \vspace{-1mm}
    \caption{{Results of (a) naive progressively increasing offsets and (b) our strategy. The naive strategy fails to produce seamless loops, causing noticeable temporal flickering between the final and first frames, as evidenced by the snow on the tree trunk indicated by the arrow.}}
    \label{fig:grouping}
    \vspace{-1mm}
\end{figure}

{Building upon Sections~\ref{sec:dependency}-\ref{sec:anchor}, we propose an \textit{Anchored Position Embedding Shifting} (Anchored Shifting) strategy to inject cyclic temporal positional information.
We begin by shifting the RoPEs across layers with progressively increasing offsets, using the same shifting operation defined in Eq.~(\ref{eq:shift_rope}).
The finding of the anchor layer and the observations in Fig.~\ref{fig:anchor_result} motivate us to keep this layer unshifted, thereby preventing artifacts while preserving semantic alignment with the user-provided prompt.
Let $l^*$ denote the index of the anchor layer.
The progressively increasing offsets are defined as\xin{:}
\begin{equation}
\begin{aligned}
    \delta_l &= (l-l^*) \bmod T', \quad l \in \{0,1,\ldots,L-1\}, \label{eq:progressive} \\
    l^* &= \arg\max_l \alpha_l,
\end{aligned}
\end{equation}
where $\delta_l$ denotes the temporal shifting offset of layer $l$, and $\arg\max$ denotes the operation used to identify the maximum degree of RoPE influence.
The second row of Fig.~\ref{fig:shifting_strategy} illustrates this design.
We visualize the DiT's temporal perception as a ring, where the gap denotes the loop boundary, i.e., the transition from the final frame to the first frame.
By assigning each layer a distinct temporal position, Eq.~(\ref{eq:progressive}) encourages the DiT to develop a global cyclic temporal perception.}

{However, as shown in Fig.~\ref{fig:grouping}, the naive strategy with progressively increasing offsets fails to produce seamless loops, resulting in noticeable temporal flickering between the final and initial frames.
This is because such a linear offset schedule implicitly assumes that all layers exert an equal influence on temporal positioning.
As discussed in Section~\ref{sec:dependency}, however, some layers have a substantially weaker effect.
Consequently, layers with negligible temporal control contribute little to perceiving and regulating transitions at the loop boundary.
The resulting insufficient accumulation of temporal control near the boundary degrades temporal continuity and leads to unstable looping behavior.}

{Correspondingly, we propose applying temporal shifts with layer-specific offsets such that each loop boundary receives nearly equal accumulated temporal control.
Based on the statistics in Fig.~\ref{fig:analysis}, we group the RoPE layers so that the control effect of RoPE within each group is approximately balanced.
By aggregating these layer-wise shifts across all layers, the DiT develops a uniformly cyclic temporal perception, thus enabling seamless looping.}

Specifically, given {attention} layers $l \in \{ 0, 1, \dots, L-1\}$, we {adaptively} assign each layer to a group 
$g_l \in \{0, 1, \dots, T'-1\}$, where $T'$ denotes the video length. 
The grouping is designed such that the total influence within each group is approximately balanced as:
\begin{equation}
\sum_{l: g_l = i} \alpha_l \approx \sum_{l: g_l = j} \alpha_l, \quad \forall i \neq j.
\end{equation}
For simplicity, we group adjacent layers whenever possible as:
\begin{equation}
g_0 \leq g_1 \leq \cdots \leq g_{L-1}.
\end{equation}
Denote the temporal shifting offset of layer $l$ as $\delta_l$, we enforce 
layers within the same group to share the same shift as:
\begin{equation}
\delta_l = s_{g_l}, \quad s_k \in \mathbb{Z}.
\end{equation}
We also set the shifting offset $\delta_{l^*}$ of the anchor layer $l^*$ to zero, i.e., $\delta_{l^*}=0$. For other groups, the shifting steps increase monotonically:
\begin{equation}
s_1 \leq s_2 \leq \cdots \leq s_{g_{l^*}} = 0 \leq \cdots \leq s_T.
\end{equation}

{A visualization of this strategy can be found in the third row of Fig.~\ref{fig:shifting_strategy}.
With grouped layer-specific offsets, temporal control is uniformly accumulated at each loop boundary position, leading to a uniform and circlic temporal perception of DiT.
As shown in Fig.~\ref{fig:grouping}, with our layer-specific shifting offset, the temporal flickering is reduced, and the video is seamlessly looping.}

\subsection{Seamless Looping Video Generation}

\begin{figure}[t!]
    \centering
    \vspace{-1mm}
    \includegraphics[width=0.99\linewidth]{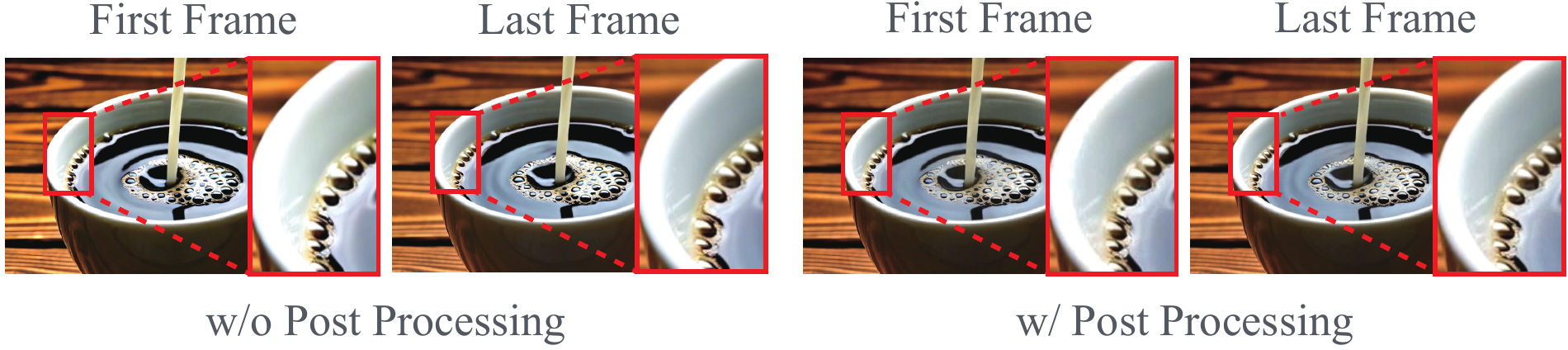}
    \vspace{-1mm}
    \caption{{Comparison of results without and with post-processing for correcting brightness and color jitter.}}
    \label{fig:post_processing}
\end{figure}

We now have a semi-training-free approach for looping video generation. We refer to it as semi-training-free due to the temporal sensitivity exhibited by the VAE decoder.
Modern video generation models typically perform diffusion-based denoising in the latent space, followed by decoding the latent representation into video via a VAE decoder.
Since recent video generation models often incorporate temporal compression in the VAE, the VAE decoder exhibits linearly temporal awareness.
{This behavior conflicts with our anchored shifting strategy, which is designed to eliminate such linearity.  
Consequently, decoding latents with the VAE may introduce minor temporal looping discontinuities, primarily manifesting as subtle variations in brightness or color, as illustrated in Fig.~\ref{fig:post_processing}.
Fortunately, these minor issues can be easily corrected with simple post-processing.}
Nevertheless, to ensure practical usability and methodological consistency, we employ minimal fine-tuning to further eliminate these artifacts.

\setlength{\tabcolsep}{2pt}
\renewcommand{\arraystretch}{1.0}
\begin{table}[t]
    \centering
    \caption{{Ablation analysis of the \emph{Proposed Anchor}, \emph{Grouped Shifting} and \emph{Finetuning}. We examine the effect of removing various components on looping video generation performance. We compare our Loopy (see the last row) with different alternative variants in terms of Cyclic Smoothness.}}
    \vspace{-2mm}
    \label{tab:component}
    \footnotesize
    \renewcommand{\arraystretch}{1.2}
    
    {\begin{tabular}{c c c||c}
    \hline

    \multirow{1}{*}{\makecell[c]{{\bf Proposed}  \textbf{$\;$Anchor}}}
    & 
    \multirow{1}{*}{\makecell[c]{{\bf Grouped}  \textbf{$\;$Shifting}}}
    &
    \multirow{1}{*}{\makecell[c]{\textbf{$\;$Fine-tuning}$\;$}}
    &\multirow{1}{*}{\makecell[c]{\textbf{Cyclic}  $\;$\textbf{Smoothness}}$\uparrow$$\;$}
   \\
\hline\hline
     &  &                  &0.9753    \\
      & &\ding{51}              &0.9802   \\
      \ding{51}&  & &0.9818\\
      \ding{51}& \ding{51} & &0.9878\\
      \ding{51}& \ding{51} & \ding{51}&\textbf{0.9925}\\

    \hline
    \end{tabular}}
    \label{tab:ablation_first}
\end{table}

\setlength{\tabcolsep}{8pt}
\renewcommand{\arraystretch}{1.2}
\begin{table*}[t]
\footnotesize
\centering
\caption{{Strategy Selection} of Loopy. We investigate the effects of anchored layer selection (\textbf{Anchored Layer}), different offset strategies (\textbf{Offset Strategy}), and fine-tuning with or without the proposed Loopy (\textbf{Fine-tuning}), and report the corresponding performance in terms of cyclic smoothness.}
\begin{tabular}{c||c c c c c||c c||c c}
\hline 
\multirow{2}{*}{\textbf{Method}} &
\multicolumn{5}{c||}{\textbf{Anchored Layer}} &
\multicolumn{2}{c||}{\textbf{Offset Strategy}} &
\multicolumn{2}{c}{\textbf{Fine-tuning}} \\
\cline{2-10}
 & \textbf{R1} & \textbf{R2} & \textbf{R3} & \textbf{R4} &
 \textbf{Proposed Anchor} & \textbf{Naive Shifting} &
 \textbf{Grouped Shifting} & \textbf{Baseline} & \textbf{Loopy} \\
\hline \hline
\textbf{Cyclic Smoothness} & 0.9779   & 0.9789 & 0.9781 & 0.9739 & \textbf{0.9878} & 0.9818 & \textbf{0.9878} & 0.9802 & \textbf{0.9925} \\
\hline
\end{tabular}
\label{tab:ablation}
\end{table*}

\begin{figure}[t!]
    \centering
    \includegraphics[width=0.99\linewidth]{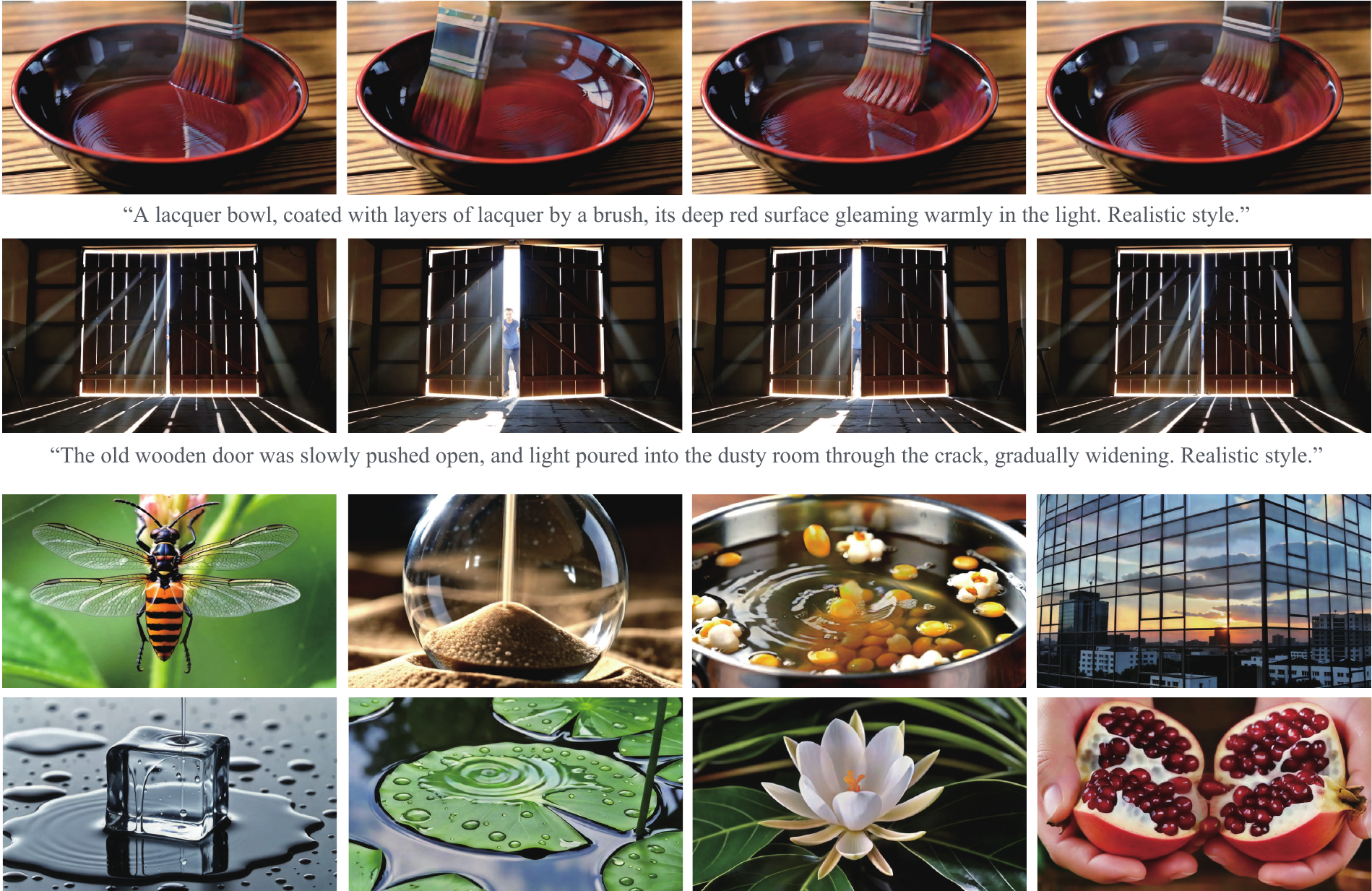}
    \vspace{-1mm}
    \caption{{Samples from our looping video training data. The first two rows demonstrate vivid motion and seamless temporal looping consistency, while the bottom two rows highlight the scene diversity of our dataset.}}
    \label{fig:dataset}
    \vspace{-1mm}
\end{figure}

We first apply {our anchored shifting to Wan 2.2 14B, generating} a small collection of looping videos. {Subsequently, Deflicker~\cite{lei2023blind} is employed to correct color inconsistencies, enhancing visual continuity. The generated videos form a high-quality looping video training dataset, with representative examples illustrated in Fig.~\ref{fig:dataset}}. The dataset contains 120 {looping} videos covering diverse scenarios, including animals, portraits, visual effects, objects, and landscapes. {Benefiting from the powerful generative capability of Wan 2.2, these videos exhibit realistic textures, coherent structures, and vivid motion variations.}

Next, we fine-tune models on this dataset using LoRA with a relatively small rank (e.g., 4). During both training and inference, DiT employs our proposed {anchored shifting strategy}, {enhances stability and promotes} semantically consistent looping video generation. In the ablation study, {we further demonstrate that removing our anchored shifting results in a substantial degradation in looping stability.}

{Beyond RGB modality}, we also extend our framework to RGBA {looping video generation}, which includes an additional transparency channel.
{We adopt} Wan-Alpha~\cite{dong2025wanalpha} as the backbone {model and apply our anchored shifting to generate RGBA looping videos, which enables} flexible content manipulation. {This is because that} RGBA looping videos are {particularly valuable} for practical content creation and editing in applications such as game development, video effects, and digital design.

\section{Experiments}
\subsection{Implementation Details}
We integrate our method with Wan2.2-T2V-A14B~\cite{wan2025wan} and Wan-Alpha~\cite{dong2025wanalpha}. We generate $120$ RGB and 120 RGBA looping videos at a resolution of $480\times 832$, comprising $53$ frames at $16$ FPS to construct our high-quality looping video dataset, using only 4 sampling steps with LightX2V~\cite{lightx2v}. Then we fine-tune Wan2.1-T2V-1.3B, Wan2.1-T2V-14B, Hunyuanvideo 1.5 and Wan-Alpha on our dataset with LoRA rank $4$. The Wan2.1-T2V-1.3B, Wan2.1-T2V-14B and Wan-Alpha were trained for $240$ steps with a batch size of $8$. The Hunyuanvideo 1.5 was trained for $4,800$ steps with a batch size of $1$. Training is conducted on 8 NVIDIA H20 GPUs. After fine-tuning, it can support videos of all resolutions and lengths that are supported by the base model.
\subsection{Evaluation Metrics}
\label{sec:metrics}
We use VBench~\cite{huang2023vbench} to assess aesthetic quality and motion smoothness of generated RGBA looping videos. Following Wan-Alpha, the VLLM model GPT-4o~\cite{gpt_4o} is used to measure text alignment, naturalness, and dynamic score. 
We also splice the second half of the video with the first half to assess cyclic smoothness using VBench. Higher scores indicate better performance. RGBA videos are rendered on a white background to ensure a fair evaluation.

\setlength{\tabcolsep}{6pt}
\begin{table}[t]

    \centering
    \caption{{Quantitative comparison with {conventional looping video generation approaches, including four pre-AI methods
    and FLF2V interpolation-based EDEN.} Our Loopy achieves state-of-the-art performance across all evaluation metrics.
    } }
    \footnotesize
    \renewcommand{\arraystretch}{1.2}
    \begin{tabular}{c||c c}
    \hline
    \multicolumn{1}{c||}{\multirow{1}*{\makecell[c]{\textbf{Method}}} }   
    &\multirow{1}{*}{\makecell[c]{\textbf{Cyclic} \textbf{Smoothness}}$\uparrow$}
    &\multirow{1}{*}{\makecell[c]{\textbf{Dynamic} \textbf{Score}}~$\uparrow$} \\
    \hline\hline
    \citet{Schodl} &0.9733 &1.74 \\
     \citet{Kwatra} &0.9751 &1.70 \\
     \citet{Agarwala} &0.9702 &1.52 \\
     \citet{liao2013automated} &0.9898 &1.30 \\
    EDEN~\cite{zhang2025eden}&0.9824&0.38 \\

    Loopy &\textbf{0.9925}&\textbf{3.20}\\ \hline
    \end{tabular}
    \label{tab:interpolation} 
\end{table}

\setlength{\tabcolsep}{12pt}
\renewcommand{\arraystretch}{1.2}
\begin{table*}[t]
    \centering
    \caption{Quantitative comparison with two currently available looping strategies, LatentMix~\cite{latentmix} and Mobius~\cite{bi2025mobius}, across four representative RGB video generation backbones: Wan 2.1 1.3B~\cite{wan2025wan}, Wan 2.1 14B~\cite{wan2025wan}, Wan 2.2 14B~\cite{wan2025wan}, and HunyuanVideo 1.5~\cite{hunyuanvideo_1.5}. Our Loopy achieves superior performance across all evaluation metrics.}
    \footnotesize
\begin{tabular}{cc||cccccc}
    \hline
     \multicolumn{2}{c||}{\multirow{2}*{\makecell[c]{\textbf{Method}}}}
    & \multirow{2}{*}{\makecell[c]{\textbf{Text} \\ \textbf{Alignment} $\uparrow$}}
    & \multirow{2}{*}{\makecell[c]{\textbf{Aesthetic} \\ \textbf{Quality} $\uparrow$}}
    & \multirow{2}{*}{\textbf{Naturalness} $\uparrow$}
    & \multirow{2}*{\makecell[c]{\textbf{Motion} \\ \textbf{Smoothness} $\uparrow$}}
    & \multirow{2}*{\makecell[c]{\textbf{Cyclic} \\ \textbf{Smoothness} $\uparrow$}}
    & \multirow{2}*{\makecell[c]{\textbf{Dynamic} \\ \textbf{Score} $\uparrow$}} \\ 
    \multicolumn{2}{c||}{} \\ \hline \hline 

    \multirow{3}{*}{Wan2.1 1.3B} 
      & LatentMix  & 3.24 & 0.5753  & 2.16 & 0.9841 & 0.9847 & 2.88 \\ 
      & Mobius  & 3.34 & 0.5803  & 2.39 & 0.9826 & 0.9826 & 2.48 \\ 
      & Ours & \textbf{3.42} & \textbf{0.6246}  & \textbf{2.56} & \textbf{0.9861} & \textbf{0.9914} & \textbf{2.92} \\ \hline

    \multirow{3}{*}{Wan2.1 14B} 
      & LatentMix  & 3.52 & 0.5664  & 2.78 & 0.9825 & 0.9876 & 2.30 \\ 
      & Mobius  & 2.78 & 0.5964  & 2.93 & 0.9718 & 0.9918 & 3.20 \\ 
      & Ours & \textbf{3.68} & \textbf{0.6474}  & \textbf{3.28} & \textbf{0.9878} & \textbf{0.9928} & \textbf{3.32} \\ \hline

    \multirow{3}{*}{Wan2.2 14B} 
      & LatentMix  & 2.86 & 0.5895  & 3.02 & 0.9777 & 0.9834 & 2.48 \\ 
      & Mobius  & 3.34  & 0.5998 & 2.87 & 0.9697 & 0.9892&2.92 \\ 
      & Ours & \textbf{3.68} & \textbf{0.6497}  & \textbf{3.39} & \textbf{0.9811} & \textbf{0.9925} & \textbf{3.20} \\ \hline

    \multirow{3}{*}{Hunyuanvideo 1.5} 
      & LatentMix  & 3.50 & 0.4950  & 3.16 & 0.9809 & 0.9752 & 3.02 \\ 
      & Mobius  & 2.60 & 0.3508  & 2.83 & 0.9722 & 0.9872 & 2.10 \\ 
      & Ours & \textbf{3.80} & \textbf{0.5146}  & \textbf{3.46} & \textbf{0.9849} & \textbf{0.9914} & \textbf{3.34} \\ \hline

    \end{tabular}
    \label{tab:compare_rgb}
\end{table*}

\subsection{{Component-wise Analysis on Loopy}}
{\emph{Proposed Anchor}, \emph{Grouped Shifting}, and \emph{Fine-tuning} are three core components of Loopy. {We examine the effect of removing various components}, with the results reported in Tab.~\ref{tab:ablation_first}. We first remove the \emph{Proposed Anchor}, \emph{Grouped Shifting}, and \emph{Fine-tuning}, {degrading our Loopy to the original baseline that is not trained on RGBA video data.}
As the original model perceives time as a linear sequence, it fails to establish temporal continuity between the final and initial frames, resulting in discontinuities that prevent seamless looping video generation (see the first row of Tab.~\ref{tab:ablation_first}).
We then fine-tune the backbone on our {constructed looping video} dataset using LoRA. {We denote this fine-tuned version as the baseline model. Without \emph{Proposed Anchor} and \emph{Grouped Shifting}, the performance gain brought by fine-tuning is limited (see the second row of Tab.~\ref{tab:ablation_first}). This is because the small-scale training data is insufficient for the model to learn the inherent cyclic temporal pattern of looping videos during large-scale pretraining.}
Next, we introduce the \emph{Proposed Anchor} and apply progressively increasing temporal offsets while keeping the anchor layer unshifted. This strategy injects cyclic temporal information while preserving the contextual prior provided by the most influential layer. Nevertheless, this strategy implicitly {assumes that all remaining layers contribute equally to temporal control}, despite their substantially different effects on temporal perception.
Consequently, temporal control is accumulated unevenly around the loop boundary, resulting in suboptimal looping performance (see the third row of Tab.~\ref{tab:ablation_first}). 
Enabling \emph{Grouped Shifting} addresses this limitation by balancing the accumulated temporal control across different temporal positions, thereby further improving cyclic smoothness (see the fourth row of Tab.~\ref{tab:ablation_first}). However, minor discontinuities may {still persist due to the linear temporal bias inherent in the temporally compressed VAE decoder.} Finally, fine-tuning the model together with the proposed shifting strategy {further mitigates} this decoder-induced mismatch. {By integrating all three components}, the complete Loopy achieves the best performance {for looping video generation (see the last row of Tab.~\ref{tab:ablation_first})}.
}
\begin{figure}[t!]
    \centering
    \includegraphics[width=0.99\linewidth]{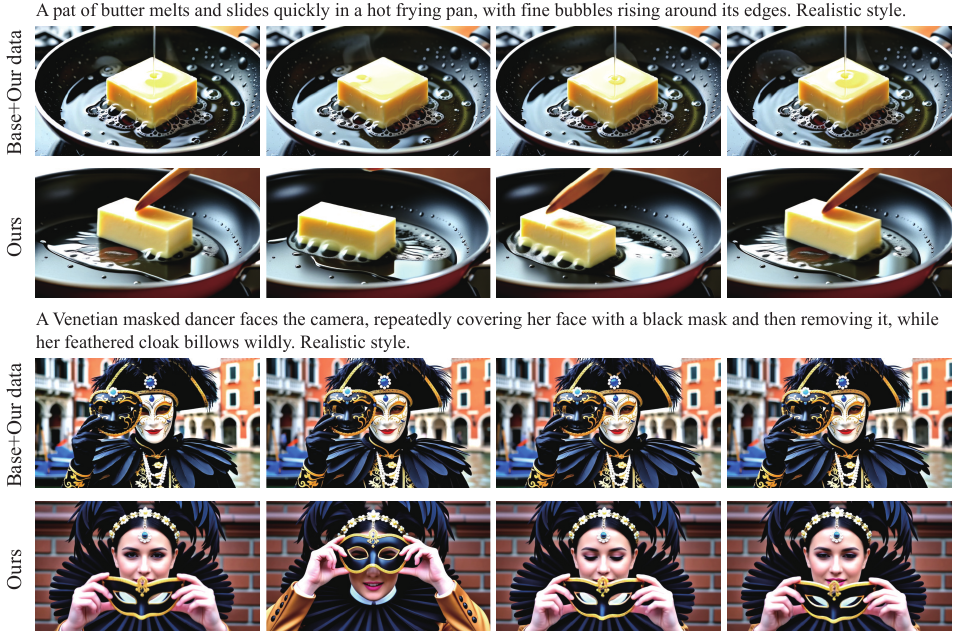}
    \vspace{-1mm}
    \caption{Ablation study of training with (Ours) and without (Base + Our Data) anchored shifting on our looping video dataset.}
    \label{fig:ablation}
    \vspace{-1mm}
\end{figure}

\subsection{{Performance Analysis on Strategy Selection}}
We first examine the effectiveness of the determination of the anchored layer in looping video generation.
This is achieved by randomly selecting four layers, excluding the one with the most pronounced RoPE control, as the anchored layer.
We remain randomly selected anchored layer unshifted, whereas the remaining layers are rotated according to our proposed anchored shifting strategy. Compared with directly applying our anchored shifting strategy to the baseline model during inference (see \textbf{Proposed Anchor} of Tab.~\ref{tab:ablation}), randomly selecting the anchored layer yields unsatisfactory performance (see \textbf{R1}-\textbf{R4}). This is because the anchored layer serves as the reference point of all RoPEs during the temporal rotation, providing contextual priors that guide the remaining layers in shaping the generated video content and mitigating artifacts. Even negligible manipulation of the anchored layer can substantially alter the generated video content, introducing artifacts or semantic inconsistencies. Hence, we define the attention layer exhibiting the most pronounced RoPE control as the anchored layer. 

We group RoPE layers to approximately balance the temporal control effect within each group. We further conduct experiments to investigate the impact of the grouping strategy and report the performance in Tab.~\ref{tab:ablation}.
Specifically, we refer to a naive strategy as progressively increasing shifting offsets. In this scenario, the temporal control effect of each RoPE within DiT is regarded as identical, yielding slight performance degradation.
This degradation arises from our observation that different RoPEs exhibit distinct control effects over temporal perception.
The naive strategy causes layers with weak RoPE control to contribute minimally to loop-boundary perception, leading to insufficient accumulated temporal control.
Consequently, the generated videos suffer from degraded temporal continuity and unstable looping.

Next, we examine the effectiveness of training with and without anchored shifting.
We denote the version without anchored shifting as the baseline model, which is simply training a LoRA on looping data.
As shown in Tab.~\ref{tab:ablation}, the baseline achieves a much lower cyclic smoothness score.
Training with anchored shifting, i.e., our final version of Loopy, achieves superior performance, as anchored shifting enables cyclic temporal perception.
The determination of the anchored layer helps generate video content with contextual coherence aligned with the text prompt (see the first case in Fig.~\ref{fig:ablation}).
By assigning layer-specific shifting offsets to each RoPE, anchored shifting enhances temporal perception at the loop boundary, enabling Loopy to generate seamless loops with diverse yet coherent motions (see the second case in Fig.~\ref{fig:ablation}).

\setlength{\tabcolsep}{14pt}
\begin{table*}[t]
    \centering
    \caption{Quantitative comparison on text-to-RGBA video generation methods. We integrate Loopy and two currently available looping strategies, LatentMix~\cite{latentmix} and Mobius~\cite{bi2025mobius}, into the RGBA video generation backbone Wan-Alpha~\cite{dong2025wanalpha}. Our Loopy achieves state-of-the-art performance across all evaluation metrics compared to competitive looping approaches. }
    \footnotesize
    \renewcommand{\arraystretch}{1.2}
    \begin{tabular}{c||c c c c c c}
    \hline
    \multicolumn{1}{c||}{\multirow{2}*{\makecell[c]{\textbf{Method}}} }   
     & \multirow{2}{*}{ \makecell[c]{\textbf{Text} \\ \textbf{Alignment}} $\uparrow$}
    & \multirow{2}{*}{\makecell[c]{\textbf{Aesthetic} \\ \textbf{Quality}}~$\uparrow$}
    & \multirow{2}{*}{{\textbf{Naturalness}$\uparrow$}}
    &\multirow{2}{*}{\makecell[c]{{\bf Motion} \\ \textbf{Smoothness}}$\uparrow$}
    &\multirow{2}{*}{\makecell[c]{\textbf{Cyclic} \\ \textbf{Smoothness}}$\uparrow$}
    &\multirow{2}{*}{\makecell[c]{\textbf{Dynamic} \\ \textbf{Score}}~$\uparrow$} \\ \\
    \hline\hline
    LatentMix + Wan-Alpha &3.22&0.576&2.65&0.9877&0.9907&3.19 \\ 
    Mobius + Wan-Alpha &3.26&0.589&2.76&0.9912 &0.9851&3.14 \\ 
    Loopy + Wan-Alpha &\textbf{3.46}&\textbf{0.636}&\textbf{3.14}&\textbf{0.9927} & \textbf{0.9921} & \textbf{3.35}\\ \hline
    \end{tabular}
    \label{tab:compare_rgba}
\end{table*}

\begin{figure*}[t!]
    \centering
    \includegraphics[width=0.99\linewidth]{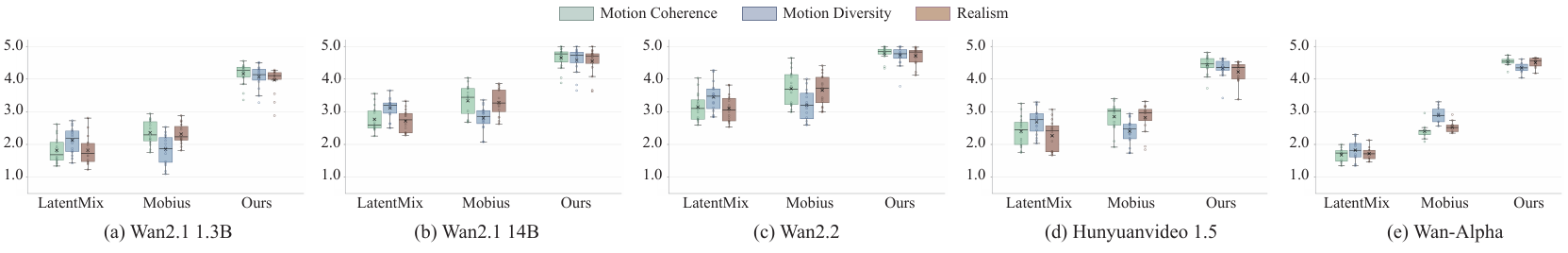}
    \vspace{-1mm}
    \caption{Box plots of the average score over the participants for each method regarding motion coherence, motion diversity, and realism.}
    \label{fig:user_study}
\end{figure*}

\begin{figure*}[t!]
    \centering
    \includegraphics[width=0.99\linewidth]{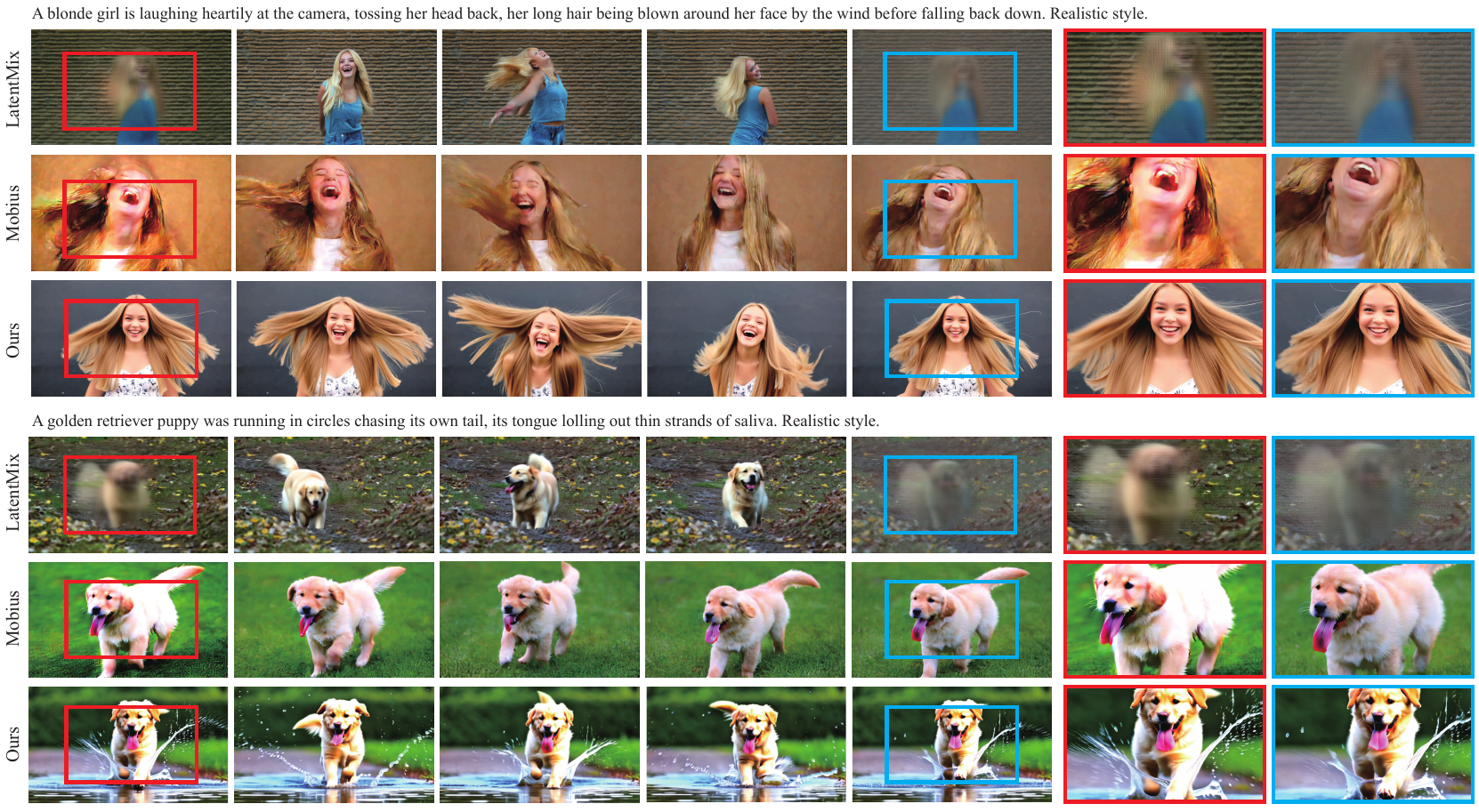}
    \vspace{-1mm}
    \caption{Comparison of different RGB looping video generation approaches integrated into Wan2.1 1.3B. {We zoom in on selected regions of the initial (see red rectangles) and final (see blue rectangles) video frames to better visualize the results produced by different approaches.}}
    \label{fig:wan1.3}
    \vspace{-1mm}
\end{figure*}

\subsection{Perceptive Evaluation Study}
We conducted an online user study to evaluate the quality of the generated RGB and RGBA looping videos, focusing on motion coherence, motion diversity, and realism. We randomly selected 100 text prompts from the test set, where RGB and RGBA video generation each comprises half of the samples. 
We utilize these inputs to generate RGB and RGBA looping videos for assessing looping video generation against two looping strategies across five backbones, including four backbones for RGB video generation and Wan-Alpha for RGBA video generation. 
We received feedback from 40 participants, including 20 males and 20 females aged between 20 and 40. Participants are required to rate each displayed video using a five-point Likert scale, where $1$ represents the worst and $5$ represents the best.
Fig.~\ref{fig:user_study} presents the comparative results on RGB and RGBA looping video generation across 3 evaluation metrics using box plots. Our Loopy receives higher user preferences than other looping strategies across all five backbones, revealing that Loopy can foster a balance between motion diversity and seamless transitions at loop boundaries.

{For each backbone/metric, we compared Loopy against each baseline using two-sided Wilcoxon signed-rank tests~\cite{wilcoxon1992individual} (N=40), with Holm–Bonferroni correction~\cite{holm1979simple} across all pairwise comparisons. Loopy significantly outperforms LatentMix/Mobius on all metrics and backbones (Holm-corrected p < 0.001; effect size r $\approx$ 0.87, indicating a large effect by Cohen's convention (r > 0.5)).
}
\begin{figure*}[!ht]
    \centering
    \vspace{-1mm}
    \includegraphics[width=0.99\linewidth]{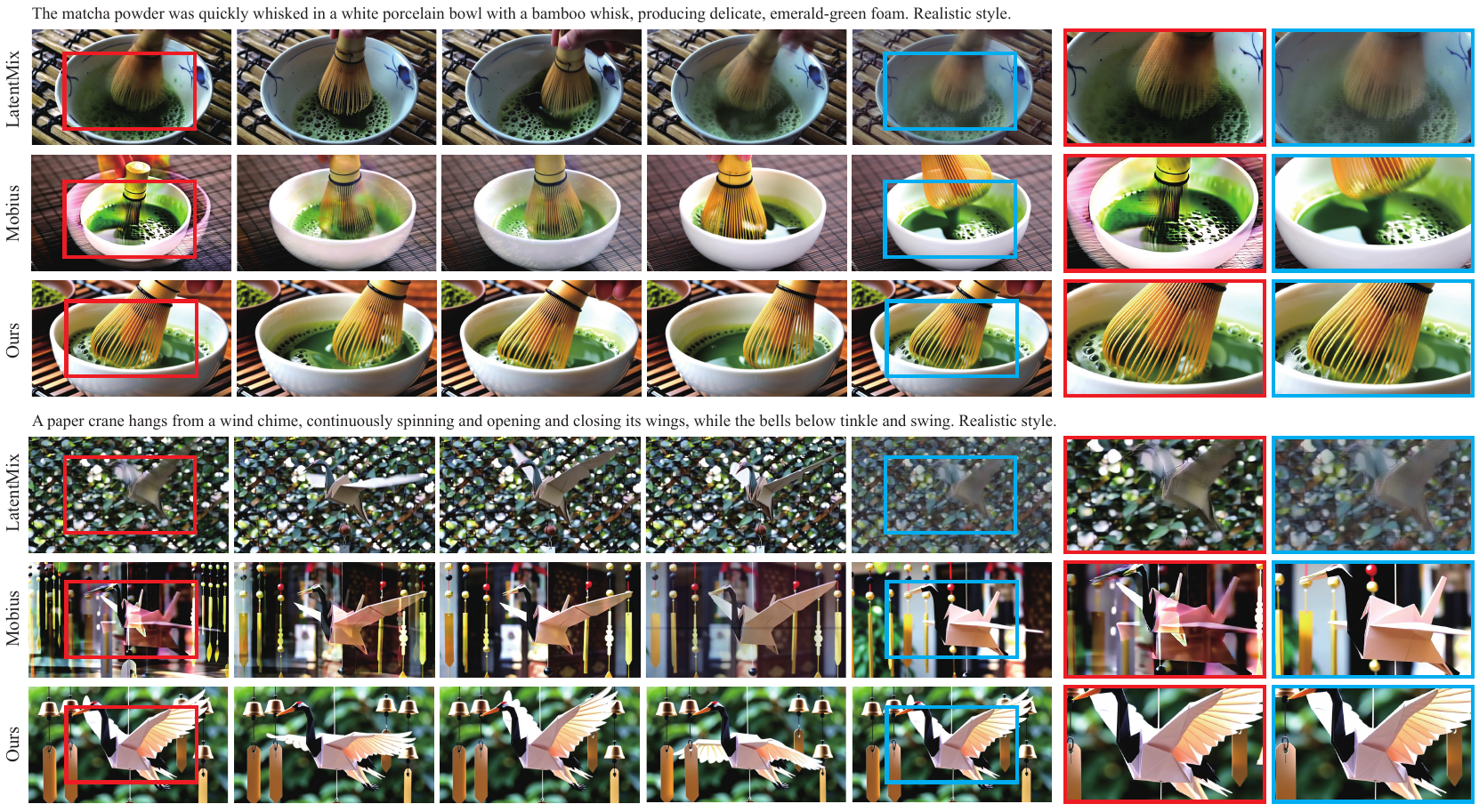}
    \vspace{-1mm}
    \caption{Comparison of different RGB looping video generation approaches integrated into Wan2.1 14B. {We zoom in on selected regions of the initial (see red rectangles) and final (see blue rectangles) video frames for better visualization. Our Loopy achieves seamless transitions at the loop boundary while mitigating artifacts.}}
    \label{fig:wan2.1}
    \vspace{-1mm}
\end{figure*}

\begin{figure*}[!ht]
    \centering
    \vspace{-1mm}
    \includegraphics[width=0.99\linewidth]{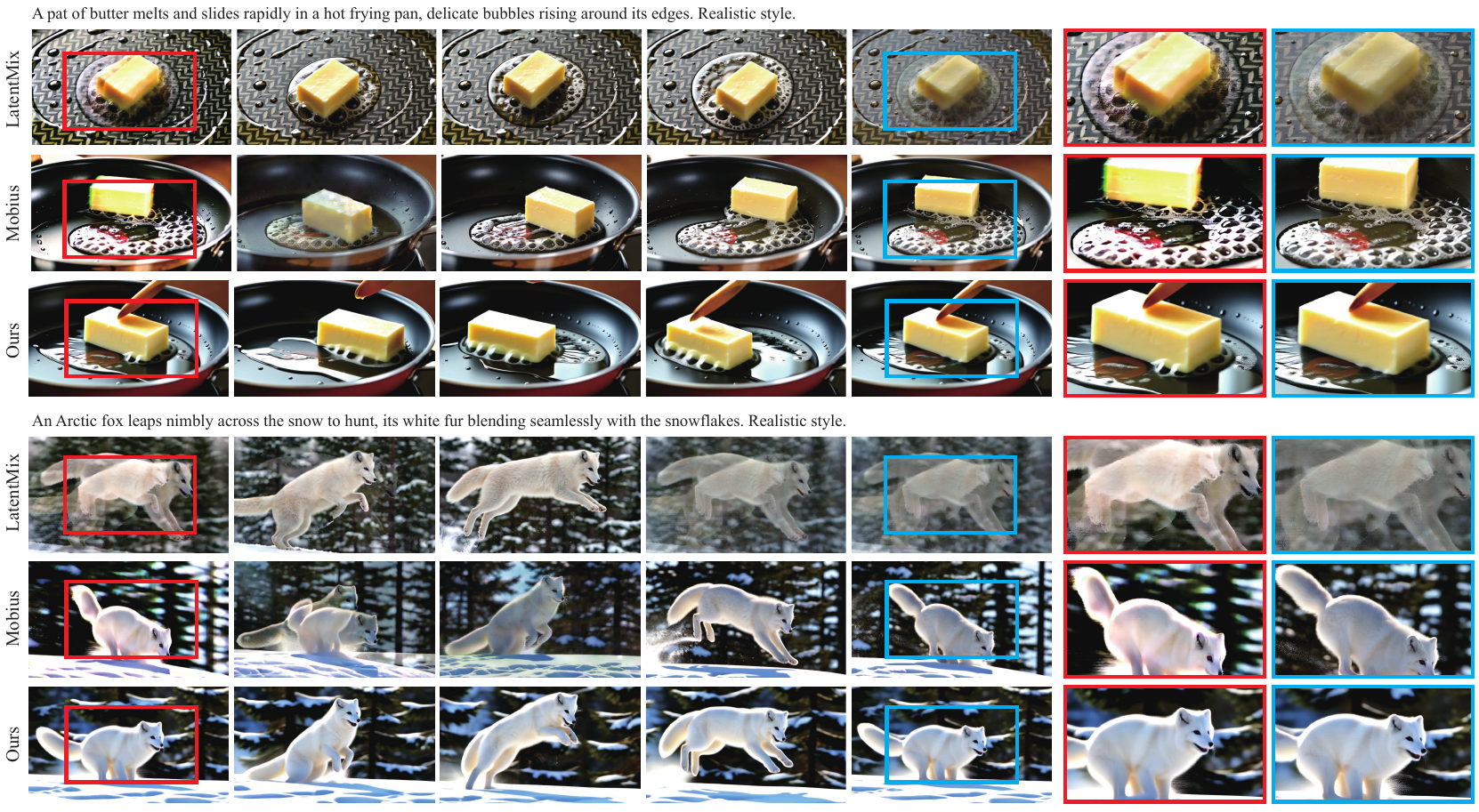}
    \vspace{-1mm}
    \caption{Comparison of different RGB looping video generation approaches integrated into Wan2.2.}
    \label{fig:wan2.2}
    \vspace{-1mm}
\end{figure*}

\subsection{State-of-the-Art Comparison}
{To provide a comprehensive evaluation, we categorize existing looping video generation approaches into two groups: (1) conventional approaches, including \citet{Schodl}, \citet{Kwatra}, \citet{Agarwala}, \citet{liao2013automated}, and the first-last-frame-to-video (FLF2V) interpolation-based EDEN~\cite{zhang2025eden}; and (2) generation-based approaches, which utilize deep generative models to generate looping videos.} 

{For the conventional approaches, we compare our Loopy against four pre-AI methods and an FLF2V interpolation-based EDEN, using two primary metrics for assessing looping quality: Dynamic Score and Cyclic Smoothness. As shown in Tab.~\ref{tab:interpolation}, our Loopy achieves the best overall performance. The pre-AI approaches construct looping videos by reusing and rearranging existing temporal content, resulting in limited motion {diversity} and unnatural looping patterns. For the interpolation-based approach, when identical start and end frames are specified, EDEN suffers from static collapse, {resulting in limited or no motion variations in the generated video loops}. Conversely, when different endpoint frames are used, {EDEN generates two video segments by reversing the start and end conditions, which are concatenated to form a video loop.} However, this strategy introduces noticeable inconsistencies between the two segments. Moreover, when the endpoint frames differ substantially, EDEN fails to produce natural and temporally coherent interpolations.}

Since {generation-based} looping video generation remains largely unexplored, {only two looping strategies are currently available }for comparison: LatentMix~\cite{latentmix} and Mobius~\cite{bi2025mobius}, both of which perform latent manipulation throughout the diffusion process. {For a comprehensive comparison}, we evaluate {our proposed Loopy and competitive approaches} using six {evaluation} metrics as described in Sec.~\ref{sec:metrics}, with quantitative results reported in {Tabs.}~\ref{tab:compare_rgb}–\ref{tab:compare_rgba}.  Across all five backbones, our method consistently outperforms LatentMix and Mobius, achieving improved text alignment, aesthetic quality, naturalness, and motion smoothness. This indicates that our approach better preserves the generation capability of each backbone, resulting in higher overall video quality.
Our {Loopy} also achieves superior cyclic smoothness, demonstrating enhanced temporal consistency between the final and initial frames and confirming the effectiveness of shifting for looping behavior. Furthermore, our approach attains higher dynamic scores, showing that the looping strategy preserves motion diversity, whereas naive latent manipulation can degrade motion fidelity.

Additionally, LatentMix, Mobius, and Loopy are all plug-and-play methods that only involve simple tensor-wise addition and multiplication operations. Therefore, they introduce negligible computational overhead and {have minimal impact on inference efficiency}.

\begin{figure*}[t!]
    \centering
    \vspace{-1mm}
    \includegraphics[width=0.99\linewidth]{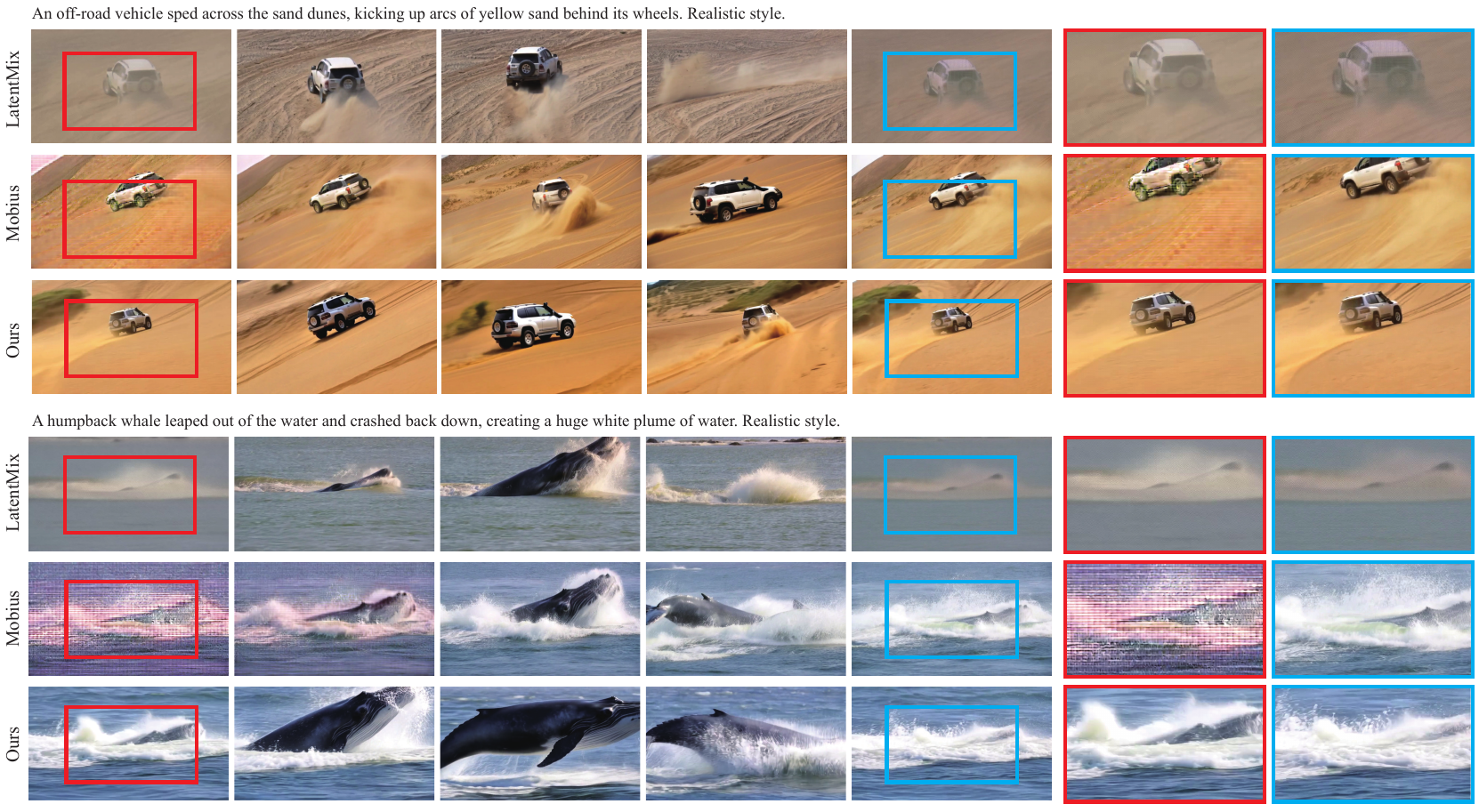}
    \vspace{-1mm}
    \caption{Comparison of different RGB looping video generation approaches integrated into Hunyuanvideo 1.5.}
    \label{fig:hunyuan}
    \vspace{-1mm}
\end{figure*}

\begin{figure*}[t!]
    \centering
    \vspace{-1mm}
    \includegraphics[width=0.99\linewidth]{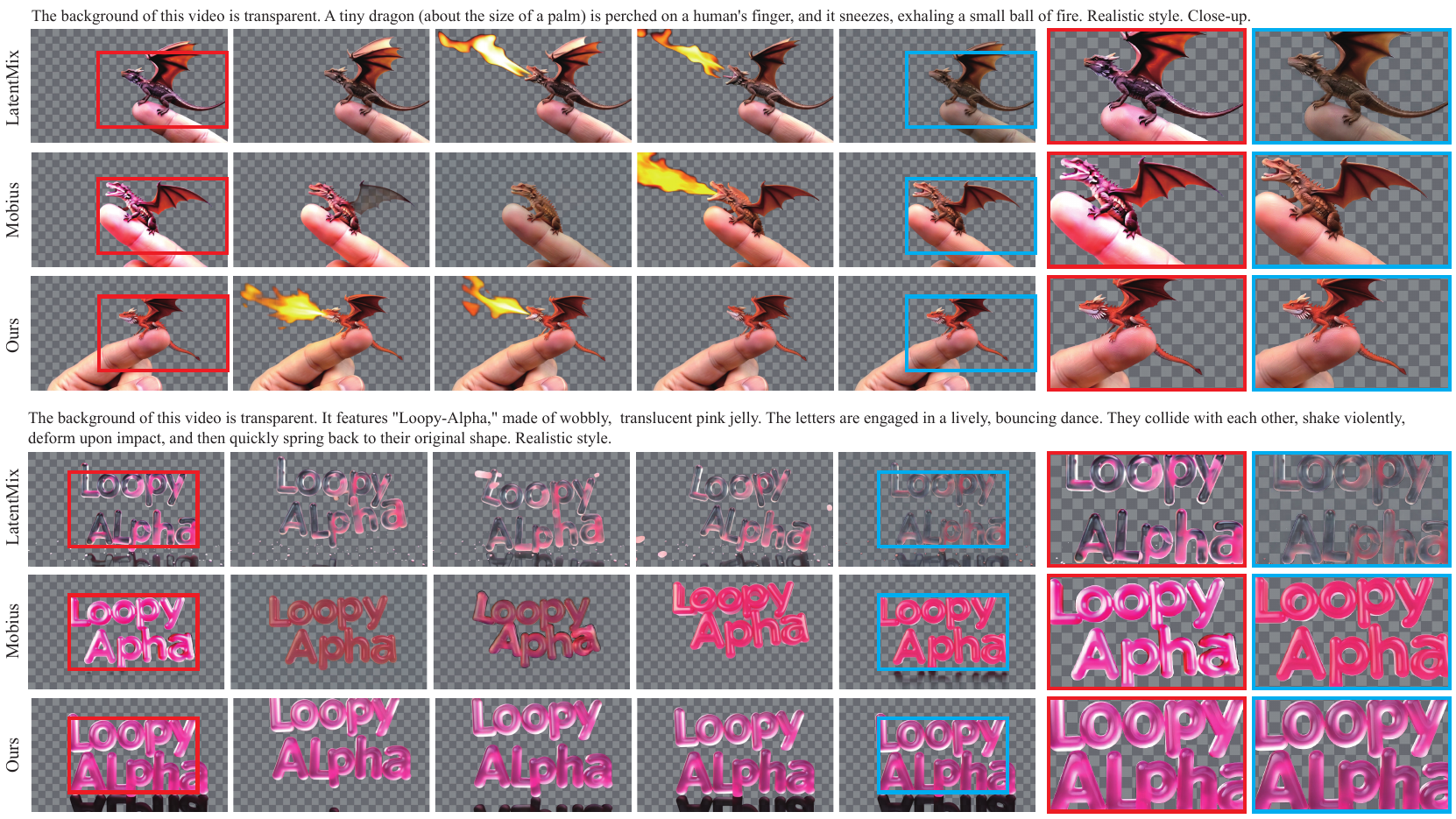}
    \vspace{-1mm}
    \caption{Comparison of different RGBA looping video generation approaches integrated into Wan-Alpha. We zoom in on selected regions of the initial (see red rectangles) and final (see blue rectangles) video frames for better visualization. Benefiting from our proposed anchored shifting strategy, Loopy achieves realistic RGBA looping video generation without introducing artifacts and incoherent color variations.}
    \label{fig:t2v}
    \vspace{1mm}
\end{figure*}

\begin{figure*}[!ht]
    \centering
    \includegraphics[width=0.99\linewidth]{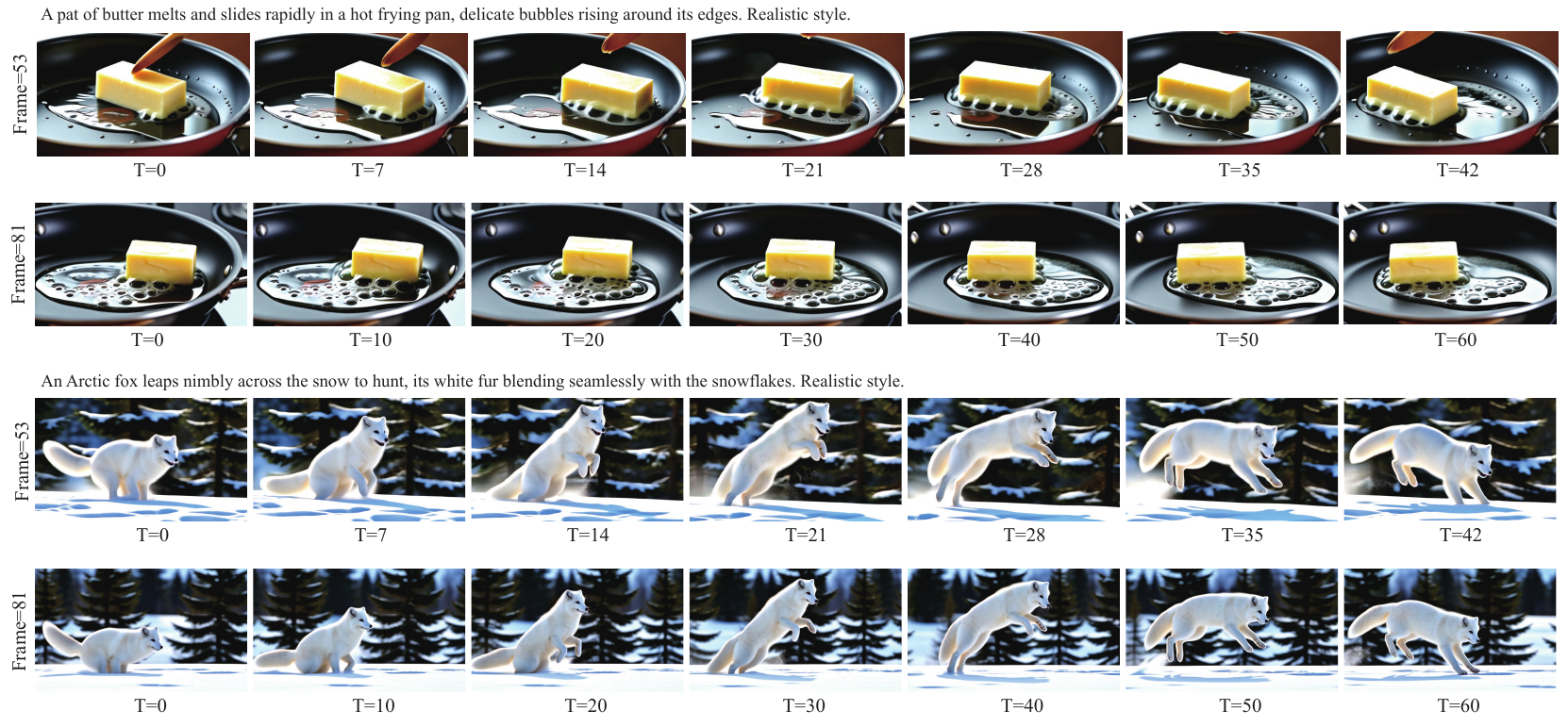}
    \vspace{-1mm}
    \caption{{Loopy can generate videos of all lengths that are supported by the base model.}}
    \label{fig:wan2.2_long}
\end{figure*}

Subjective comparisons are shown in Figs.~\ref{fig:wan1.3},~\ref{fig:wan2.1},~\ref{fig:wan2.2},~\ref{fig:hunyuan} and ~\ref{fig:t2v}.

\begin{figure}[t!]
    \centering
    \vspace{-1mm}
    \includegraphics[width=0.99\linewidth]{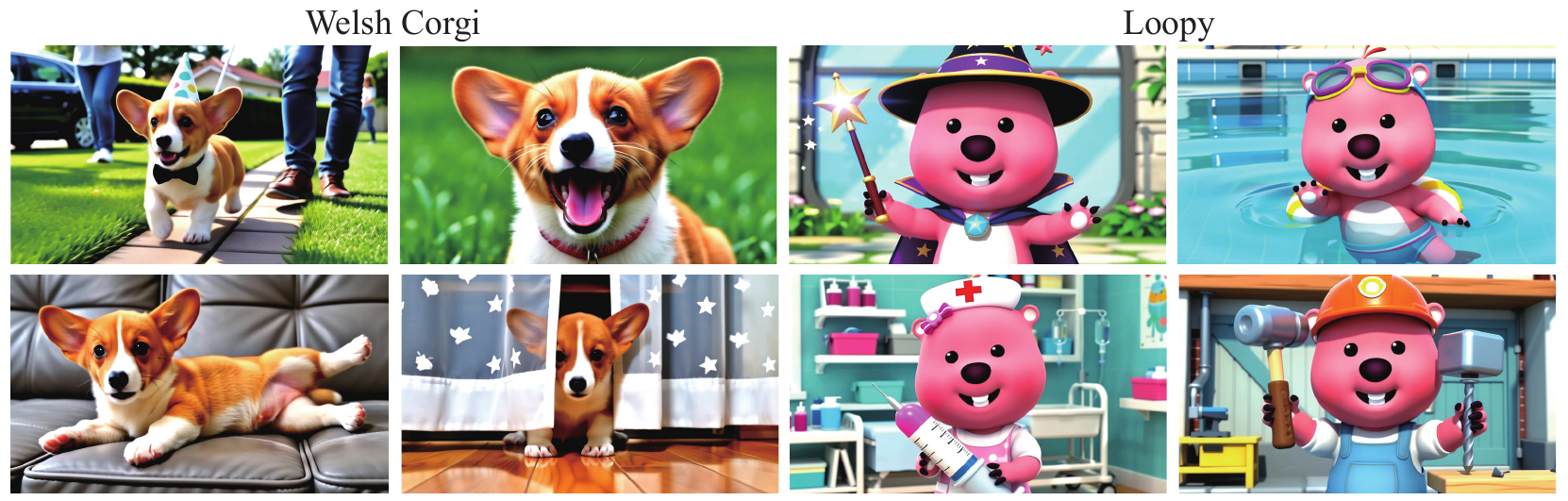}
    \vspace{-1mm}
    \caption{Application of our Loopy for character control. It can generate stable Welsh Corgi (left) and Loopy (right) characters across various scenes.}
    \label{fig:loopy}
    \vspace{-1mm}
\end{figure}

\begin{figure}[t!]
    \centering
    \vspace{-1mm}
    \includegraphics[width=0.99\linewidth]{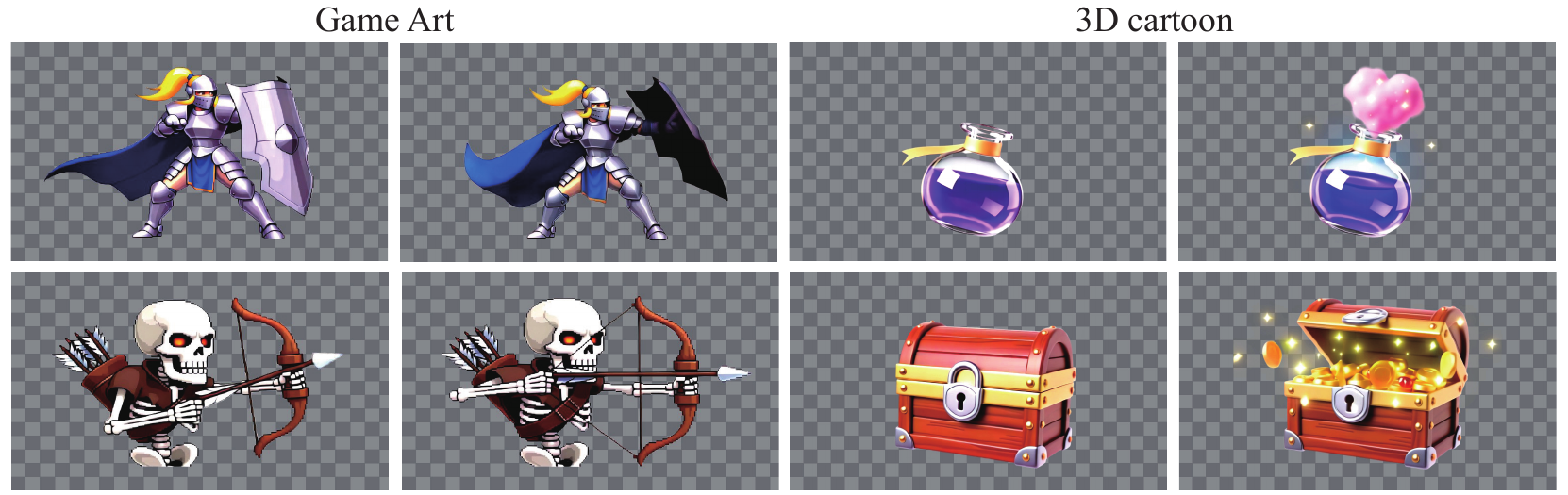}
    \vspace{-1mm}
    \caption{Application of our Loopy for style control. It can generate videos in Game Art (left) and 3D Cartoon (right) styles with different content.}
    \label{fig:style}
    \vspace{-1mm}
\end{figure}

\begin{figure}[t!]
    \centering
    \vspace{-1mm}
    \includegraphics[width=0.99\linewidth]{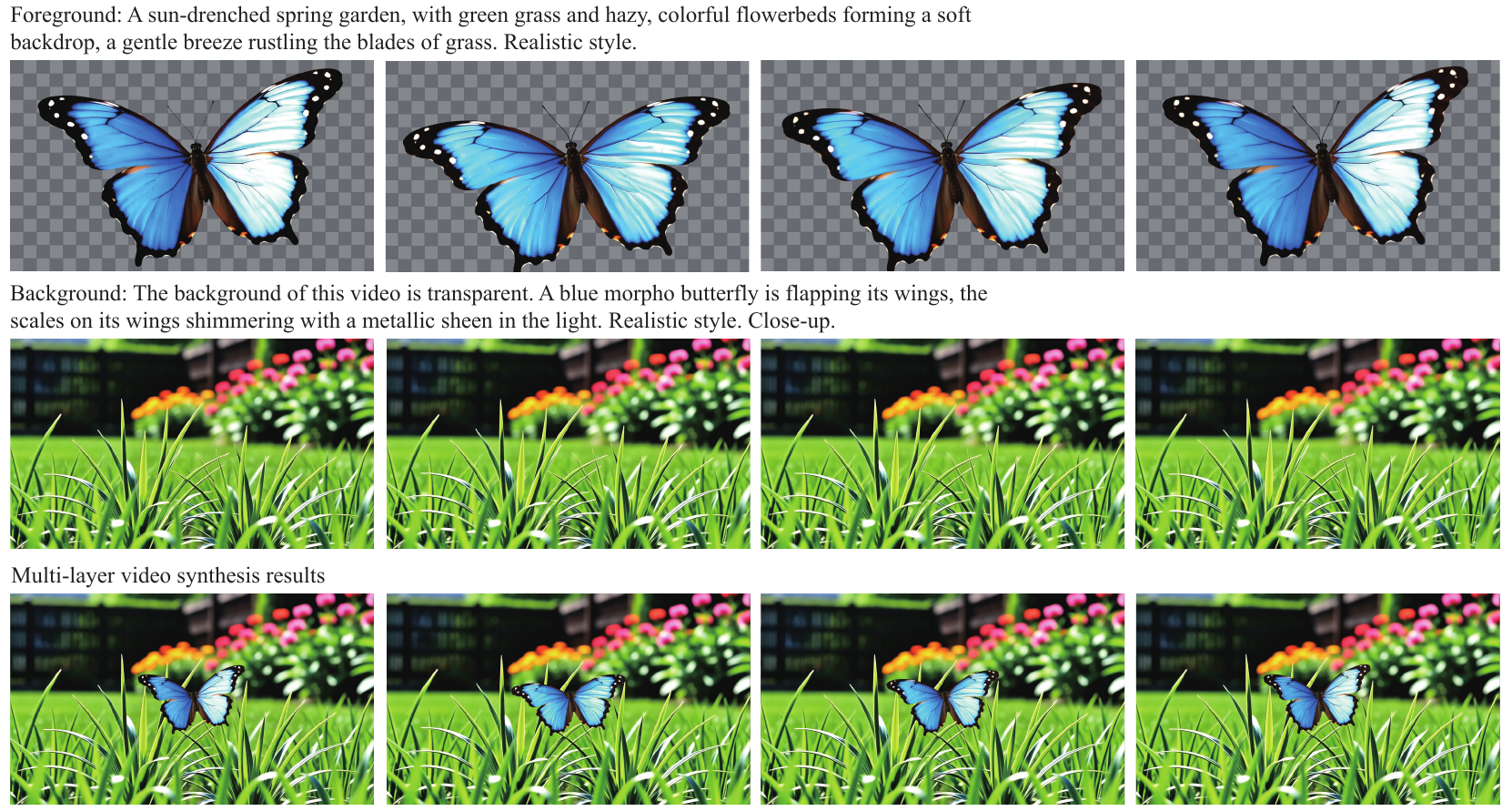}
    \vspace{-1mm}
    \caption{Application of our Loopy for multi-layer video generation. By combining the generated RGB and RGBA videos, the model can produce a vivid multi-layer looping video.}
    \label{fig:multi}
    \vspace{-1mm}
\end{figure}

\subsection{Applications}
{Our method can support any lengths that are supported by the base model. In Fig.~\ref{fig:wan2.2_long}, we provide the 53- and 81-frame results.} Fig.~\ref{fig:teaser}, Fig.~\ref{fig:loopy}, Fig.~\ref{fig:style}, and Fig.~\ref{fig:multi} showcase the versatility of Loopy across various practical applications, including character control, style control, and multi-layer looping video generation. Beyond these examples, Loopy also holds great potential for applications such as live wallpapers, web graphics, game visuals, and social media content.

\section{Conclusion and Discussion}
In this paper, we are the first to analyze RoPE's temporal control at different attention layers within DiT. Built upon two core findings -- \textit{different RoPEs exhibit distinct control effects over temporal perception} and \textit{the attention layer exhibiting the most pronounced RoPE control effect acts as an anchor} -- we propose \textit{Anchored Position Embedding Shifting} (Anchored Shifting) that assigns layer-specific temporal shifting offset to each RoPE for preserving temporal coherence and enabling seamless loop-boundary transition. We demonstrate the practicality of Loopy through various applications, ranging from RGB/RGBA looping video generation to multi-layer looping video generation. These capabilities make Loopy a versatile and robust tool for downstream tasks in both the AIGC community and broader industrial applications, including live wallpapers, web graphics, advertising design, and visual effects.

Our Loopy supports the generation of multi-layer videos without the consideration of illumination or shadow consistency, which may lead to inconsistencies in the generated video content. This limitation could potentially be addressed by introducing shared attention mechanisms between foreground and background DiTs, although such an investigation is beyond the scope of this paper. In our future work, we plan to explore more efficient solutions for realistic multi-layer looping video generation.

\section*{Acknowledgements}
We thank the anonymous reviewers for their constructive comments. This work was supported by National Natural Science Foundation of China (No.62476192).

\bibliographystyle{ACM-Reference-Format}
\bibliography{sample-bibliography}

@article{LayerDiffuse,
  author       = {Lvmin Zhang and
                  Maneesh Agrawala},
  title        = {Transparent Image Layer Diffusion using Latent Transparency},
  journal      = {ACM TOG},
  year         = {2024},
}

@article{su2024roformer,
  title={Roformer: Enhanced transformer with rotary position embedding},
  author={Su, Jianlin and Ahmed, Murtadha and Lu, Yu and Pan, Shengfeng and Bo, Wen and Liu, Yunfeng},
  journal={Neurocomputing},
  year={2024},
}

@inproceedings{peebles2023scalable,
  title={Scalable diffusion models with transformers},
  author={Peebles, William and Xie, Saining},
  booktitle={IEEE ICCV},
  year={2023}
}

@article{chen2025transanimate,
      title={TransAnimate: Taming Layer Diffusion to Generate RGBA Video}, 
      author={Xuewei Chen and Zhimin Chen and Yiren Song},
      year={2025},
      journal={ArXiv preprint},
}

@inproceedings{TransVDM,
  author       = {Menghao Li and
                  Zhenghao Zhang and
                  Junchao Liao and
                  Long Qin and
                  Weizhi Wang},
  title        = {TransVDM: Motion-Constrained Video Diffusion Model for Transparent
                  Video Synthesis},
  booktitle    = {ICASSP},
  year         = {2025},
}

@inproceedings{layeranimate,
  author       = {Jingqi Bai and
                  Jingkai Zhou and
                  Benzhi Wang and
                  Weihua Chen and
                  Yang Yang and
                  Zhen Lei and
                  Fan Wang},
  title        = {Layer-Animate for Transparent Video Generation},
  booktitle    = {ICASSP},
  year         = {2025},
}

@inproceedings{ILDiff,
  author       = {Ting Zhang and
                  Zhiqiang Yuan and
                  Yeshuang Zhu and
                  Jie Zhou and
                  Jinchao Zhang},
  title        = {ILDiff: Generate Transparent Animated Stickers by Implicit Layout Distillation},
  booktitle    = {ICASSP},
  year         = {2025},
}

@inproceedings{dong2025wanalpha,
  title={Video generation with stable transparency via shiftable rgb-a distribution learner},
  author={Dong, Haotian and Wang, Wenjing and Li, Chen and Lyu, Jing and Lin, Di},
  booktitle={IEEE CVPR},
  year={2026}
}

@inproceedings{wang2025transpixeler,
  title={TransPixeler: Advancing Text-to-Video Generation with Transparency},
  author={Wang, Luozhou and Li, Yijun and Chen, Zhifei and Wang, Jui-Hsien and Zhang, Zhifei and Zhang, He and Lin, Zhe and Chen, Ying-Cong},
  booktitle={IEEE CVPR},
  year={2025}
}

@inproceedings{bi2025mobius,
  title={Mobius: Text to Seamless Looping Video Generation via Latent Shift},
  author={Bi, Xiuli and Yuan, Jianfei and Liu, Bo and Zhang, Yong and Cun, Xiaodong and Pun, Chi-Man and Xiao, Bin},
  booktitle={ACM SIGGRAPH},
  year={2025}
}

@article{halperin2021endless,
  title={Endless loops: detecting and animating periodic patterns in still images},
  author={Halperin, Tavi and Hakim, Hanit and Vantzos, Orestis and Hochman, Gershon and Benaim, Netai and Sassy, Lior and Kupchik, Michael and Bibi, Ofir and Fried, Ohad},
  journal={ACM TOG},
  year={2021},
}

@article{mahapatra2023text,
  title={Text-guided synthesis of eulerian cinemagraphs},
  author={Mahapatra, Aniruddha and Siarohin, Aliaksandr and Lee, Hsin-Ying and Tulyakov, Sergey and Zhu, Jun-Yan},
  journal={ACM TOG},
  year={2023},
}

@inproceedings{bertiche2023blowing,
  title={Blowing in the wind: Cyclenet for human cinemagraphs from still images},
  author={Bertiche, Hugo and Mitra, Niloy J and Kulkarni, Kuldeep and Huang, Chun-Hao P and Wang, Tuanfeng Y and Madadi, Meysam and Escalera, Sergio and Ceylan, Duygu},
  booktitle={IEEE CVPR},
  year={2023}
}

@article{mahapatra2026dreamloop,
  title={DreamLoop: Controllable Cinemagraph Generation from a Single Photograph},
  author={Mahapatra, Aniruddha and Mai, Long and Ham, Cusuh and Liu, Feng},
  journal={ArXiv preprint},
  year={2026}
}

@inproceedings{wang2024loopanimate,
  title={Loopanimate: Loopable salient object animation},
  author={Wang, Fanyi and Liu, Peng and Hu, Haotian and Meng, Dan and Su, Jingwen and Xu, Jinjin and Zhang, Yanhao and Ren, Xiaoming and Zhang, Zhiwang},
  booktitle={ACMMM Asia},
  year={2024}
}

@article{wan2025wan,
  title={Wan: Open and advanced large-scale video generative models},
  author={{Wan Team}},
  journal={ArXiv preprint},
  year={2025}
}

@InProceedings{huang2023vbench,
     title={{VBench}: Comprehensive Benchmark Suite for Video Generative Models},
     author={Huang, Ziqi and He, Yinan and Yu, Jiashuo and Zhang, Fan and Si, Chenyang and Jiang, Yuming and Zhang, Yuanhan and Wu, Tianxing and Jin, Qingyang and Chanpaisit, Nattapol and Wang, Yaohui and Chen, Xinyuan and Wang, Limin and Lin, Dahua and Qiao, Yu and Liu, Ziwei},
     booktitle={IEEE CVPR},
     year={2024}
 }

@article{gpt_4o,
  author       = {OpenAI},
  title        = {{GPT-4} Technical Report},
    journal = {ArXiv preprint},
  year         = {2023},
}

@misc{latentmix,
  author       = {dribnet},
  title        = {CogVideo (with\_looping branch)},
  year         = {2024},
  howpublished = {\url{https://github.com/dribnet/CogVideo/tree/with_looping}},
 publisher = {GitHub},
 journal = {GitHub repository},
}

@misc{lightx2v,
 author = {LightX2V Contributors},
 title = {LightX2V: Light Video Generation Inference Framework},
 year = {2025},
 publisher = {GitHub},
 journal = {GitHub repository},
 howpublished = {\url{https://github.com/ModelTC/lightx2v}},
}

@article{HunyuanVideo,
  author       = {Weijie Kong and
                  Qi Tian and
                  Zijian Zhang and
                  Rox Min and
                  Zuozhuo Dai and
                  Jin Zhou and
                  Jiangfeng Xiong and
                  Xin Li and
                  Bo Wu and
                  Jianwei Zhang and
                  Kathrina Wu and
                  Qin Lin and
                  Junkun Yuan and
                  Yanxin Long and
                  Aladdin Wang and
                  Andong Wang and
                  Changlin Li and
                  Duojun Huang and
                  Fang Yang and
                  Hao Tan and
                  Hongmei Wang and
                  Jacob Song and
                  Jiawang Bai and
                  Jianbing Wu and
                  Jinbao Xue and
                  Joey Wang and
                  Kai Wang and
                  Mengyang Liu and
                  Pengyu Li and
                  Shuai Li and
                  Weiyan Wang and
                  Wenqing Yu and
                  Xinchi Deng and
                  Yang Li and
                  Yi Chen and
                  Yutao Cui and
                  Yuanbo Peng and
                  Zhentao Yu and
                  Zhiyu He and
                  Zhiyong Xu and
                  Zixiang Zhou and
                  Zunnan Xu and
                  Yangyu Tao and
                  Qinglin Lu and
                  Songtao Liu and
                  Daquan Zhou and
                  Hongfa Wang and
                  Yong Yang and
                  Di Wang and
                  Yuhong Liu and
                  Jie Jiang and
                  Caesar Zhong},
  title        = {HunyuanVideo: {A} Systematic Framework For Large Video Generative
                  Models},
  journal      = {ArXiv preprint},
  year         = {2024},
}

@article{hunyuanvideo_1.5,
      title={HunyuanVideo 1.5 Technical Report}, 
      author={{Hunyuan Team}},
  journal      = {ArXiv preprint},
      year={2025},
}

@article{seedance15pronative,
      title={Seedance 1.5 pro: A Native Audio-Visual Joint Generation Foundation Model}, 
      author={{Seedance Team}},
  journal      = {ArXiv preprint},
      year={2025},
}

@article{show_1,
  author       = {David Junhao Zhang and
                  Jay Zhangjie Wu and
                  Jia{-}Wei Liu and
                  Rui Zhao and
                  Lingmin Ran and
                  Yuchao Gu and
                  Difei Gao and
                  Mike Zheng Shou},
  title        = {Show-1: Marrying Pixel and Latent Diffusion Models for Text-to-Video
                  Generation},
  journal      = {IJCV},
  year         = {2025},
}

@inproceedings{Align_Your_Latents,
  author       = {Andreas Blattmann and
                  Robin Rombach and
                  Huan Ling and
                  Tim Dockhorn and
                  Seung Wook Kim and
                  Sanja Fidler and
                  Karsten Kreis},
  title        = {Align Your Latents: High-Resolution Video Synthesis with Latent Diffusion
                  Models},
  booktitle    = {IEEE CVPR},
  year         = {2023},
}

@article{open_sora_2,
  author       = {Xiangyu Peng and
                  Zangwei Zheng and
                  Chenhui Shen and
                  Tom Young and
                  Xinying Guo and
                  Binluo Wang and
                  Hang Xu and
                  Hongxin Liu and
                  Mingyan Jiang and
                  Wenjun Li and
                  Yuhui Wang and
                  Anbang Ye and
                  Gang Ren and
                  Qianran Ma and
                  Wanying Liang and
                  Xiang Lian and
                  Xiwen Wu and
                  Yuting Zhong and
                  Zhuangyan Li and
                  Chaoyu Gong and
                  Guojun Lei and
                  Leijun Cheng and
                  Limin Zhang and
                  Minghao Li and
                  Ruijie Zhang and
                  Silan Hu and
                  Shijie Huang and
                  Xiaokang Wang and
                  Yuanheng Zhao and
                  Yuqi Wang and
                  Ziang Wei and
                  Yang You},
  title        = {Open-Sora 2.0: Training a Commercial-Level Video Generation Model
                  in {\textdollar}200k},
  journal      = {ArXiv preprint},
  year         = {2025},
}

@article{klingteam2025klingomnitechnicalreport,
      title={Kling-Omni Technical Report}, 
      author={{Kling Team}},
      year={2025},
  journal      = {ArXiv preprint},
}

@article{seedance2026seedance20advancingvideo,
      title={Seedance 2.0: Advancing Video Generation for World Complexity}, 
      author={{Seedance Team}},
      year={2026},
  journal      = {ArXiv preprint},
}

@inproceedings{ho2020denoising,
  author       = {Jonathan Ho and
                  Ajay Jain and
                  Pieter Abbeel},
  title        = {Denoising Diffusion Probabilistic Models},
  booktitle    = {NeurIPS},
  year         = {2020},
}

@inproceedings{Phenaki,
  author       = {Ruben Villegas and
                  Mohammad Babaeizadeh and
                  Pieter{-}Jan Kindermans and
                  Hernan Moraldo and
                  Han Zhang and
                  Mohammad Taghi Saffar and
                  Santiago Castro and
                  Julius Kunze and
                  Dumitru Erhan},
  title        = {Phenaki: Variable Length Video Generation from Open Domain Textual
                  Descriptions},
  booktitle    = {ICLR},
  year         = {2023},}

@inproceedings{lipman2023flowmatching,
  author       = {Yaron Lipman and
                  Ricky T. Q. Chen and
                  Heli Ben{-}Hamu and
                  Maximilian Nickel and
                  Matthew Le},
  title        = {Flow Matching for Generative Modeling},
  booktitle    = {ICLR},
  year         = {2023},
}

@inproceedings{sit,
  author       = {Nanye Ma and
                  Mark Goldstein and
                  Michael S. Albergo and
                  Nicholas M. Boffi and
                  Eric Vanden{-}Eijnden and
                  Saining Xie},
  title        = {SiT: Exploring Flow and Diffusion-Based Generative Models with Scalable
                  Interpolant Transformers},
  booktitle    = {ECCV},
  year         = {2024},
}

@inproceedings{sd3,
  author       = {Patrick Esser and
                  Sumith Kulal and
                  Andreas Blattmann and
                  Rahim Entezari and
                  Jonas M{\"{u}}ller and
                  Harry Saini and
                  Yam Levi and
                  Dominik Lorenz and
                  Axel Sauer and
                  Frederic Boesel and
                  Dustin Podell and
                  Tim Dockhorn and
                  Zion English and
                  Robin Rombach},
  title        = {Scaling Rectified Flow Transformers for High-Resolution Image Synthesis},
  booktitle    = {ICML},
  year         = {2024},
}

@inproceedings{unet,
  author       = {Olaf Ronneberger and
                  Philipp Fischer and
                  Thomas Brox},
  title        = {U-Net: Convolutional Networks for Biomedical Image Segmentation},
  booktitle    = {MICCAI},
  year         = {2015},
}

@inproceedings{ldm,
  author       = {Robin Rombach and
                  Andreas Blattmann and
                  Dominik Lorenz and
                  Patrick Esser and
                  Bj{\"{o}}rn Ommer},
  title        = {High-Resolution Image Synthesis with Latent Diffusion Models},
  booktitle    = {IEEE CVPR},
  year         = {2022},
}

@inproceedings{lei2023blind,
  title={Blind video deflickering by neural filtering with a flawed atlas},
  author={Lei, Chenyang and Ren, Xuanchi and Zhang, Zhaoxiang and Chen, Qifeng},
  booktitle={IEEE CVPR},
  year={2023}
}

@inproceedings{transformer_2017,
  author       = {Ashish Vaswani and
                  Noam Shazeer and
                  Niki Parmar and
                  Jakob Uszkoreit and
                  Llion Jones and
                  Aidan N. Gomez and
                  Lukasz Kaiser and
                  Illia Polosukhin},
  title        = {Attention is All you Need},
  booktitle    = {NeurIPS},
  year         = {2017},
}

@article{liao2013automated,
  title={Automated video looping with progressive dynamism},
  author={Liao, Zicheng and Joshi, Neel and Hoppe, Hugues},
  journal={ACM TOG},
  year={2013},
}

@inproceedings{Schodl,
author = {Sch\"{o}dl, Arno and Szeliski, Richard and Salesin, David H. and Essa, Irfan},
title = {Video textures},
year = {2000},
booktitle = {ACM SIGGRAPH},
}

@article{Kwatra,
author = {Kwatra, Vivek and Sch\"{o}dl, Arno and Essa, Irfan and Turk, Greg and Bobick, Aaron},
title = {Graphcut textures: image and video synthesis using graph cuts},
year = {2003},
journal = {ACM TOG},
}

@inproceedings{Agarwala,
author = {Agarwala, Aseem and Zheng, Ke Colin and Pal, Chris and Agrawala, Maneesh and Cohen, Michael and Curless, Brian and Salesin, David and Szeliski, Richard},
title = {Panoramic video textures},
year = {2005},
booktitle = {ACM SIGGRAPH},
}

@inproceedings{zhang2025eden,
  title={Enhanced Diffusion for High-quality Large-motion Video Frame Interpolation},
  author={Zhang, Zihao and Chen, Haoran and Zhao, Haoyu and Lu, Guansong and Fu, Yanwei and Xu, Hang and Wu, Zuxuan},
  booktitle={IEEE CVPR},
  year={2025}
}

@inproceedings{wilcoxon1992individual,
  title={Individual comparisons by ranking methods},
  author={Wilcoxon, Frank},
  booktitle={Breakthroughs in Statistics: Methodology and Distribution},
  year={1992},
}

@article{holm1979simple,
  title={A simple sequentially rejective multiple test procedure},
  author={Holm, Sture},
  journal={Scandinavian Journal of Statistics},
  year={1979},
}

@article{wang2025hrc,
  title={HRC-Net: Learning Visual Hypothesis, Representative, and Collaboration for Multi-Domain Image Inpainting},
  author={Wang, Xin and Lin, Di and Su, Wanchao and Du, Ji and Zhang, Renjie and Zhang, Jie and Dong, Haotian and Xu, Ke and Guo, Qing and Li, Ping},
  journal={ACM TOG},
  year={2025},
}

@article{wu2026x2hdr,
  title={X2HDR: HDR Image Generation in a Perceptually Uniform Space},
  author={Wu, Ronghuan and Su, Wanchao and Ma, Kede and Liao, Jing and Mantiuk, Rafa{\l} K},
  journal={arXiv preprint arXiv:2602.04814},
  year={2026}
}

@INPROCEEDINGS{10377659,
  author={Xu, Ke and Hancke, Gerhard Petrus and Lau, Rynson W. H.},
  booktitle={IEEE ICCV}, 
  title={Learning Image Harmonization in the Linear Color Space}, 
  year={2023},
}

@INPROCEEDINGS{11444877,
  author={Li, Yiyu and Wang, Haoyuan and Xu, Ke and Hancke, Gerhard Petrus and Lau, Rynson W.H.},
  booktitle={IEEE ICCV}, 
  title={SeHDR: Single-Exposure HDR Novel View Synthesis Via 3D Gaussian Bracketing}, 
  year={2025},
}

@inproceedings{10.1145/3757377.3763929,
author = {Qu, Zefan and Wang, Zhenwei and Wang, Haoyuan and Xu, Ke and Hancke, Gerhard Petrus and Lau, Rynson. W. H.},
title = {StyleSculptor: Zero-Shot Style-Controllable 3D Asset Generation with Texture-Geometry Dual Guidance},
year = {2025},
booktitle = {ACM SIGGRAPH Asia},
}
\end{document}